\documentclass[Journal,letterpaper]{ascelike-new}
\usepackage[utf8]{inputenc}
\usepackage[T1]{fontenc}
\usepackage{lmodern}
\usepackage{graphicx}
\usepackage[style=base,figurename=Fig.,labelfont=bf,labelsep=period]{caption}
\usepackage{subcaption}
\usepackage{amsmath}
\usepackage{newtxtext,newtxmath}
\usepackage[colorlinks=true,citecolor=red,linkcolor=black]{hyperref}

\usepackage{subcaption}
\usepackage{algorithm}
\usepackage{algpseudocode}
\usepackage{setspace}

\begin{document}
\nolinenumbers
\title{Image-Guided Pavement Defect Recognition in GPR Data with novel 3D Deep Learning Architecture}

\author[1]{Yuandong Pan, Ph.D.}
\author[2]{Linjun Lu, Ph.D.}
\author[3]{Mudan Wang, Ph.D.}
\author[4]{Florian Noichl}
\author[5]{Fan Xue, Ph.D.}
\author[6]{Brian Sheil, Ph.D.}
\author[7]{Lavindra de Silva, Ph.D.}
\author[8]{André Borrmann, Ph.D.}
\author[9]{Ioannis Brilakis, Ph.D.}

\affil[1]{Postdoctoral scholar, School of Engineering, Stanford University, 473 Via Ortega, Stanford}
\affil[2]{Research fellow, Department of Engineering, University of Cambridge, 7a JJ Thomson Avenue, Cambridge. Corresponding Email: ll718@cam.ac.uk}
\affil[3]{Assistant research professor, Department of Engineering, University of Cambridge, 7a JJ Thomson Avenue, Cambridge}
\affil[4]{Postdoctoral Scholar, School of Engineering and Design, Technical University of Munich, Arcisstrasse 21, Munich}
\affil[5]{Associate professor, Department of Real Estate and Construction, University of Hong Kong, Pok Fu Lam Rd, Hong Kong}
\affil[6]{Associate professor, Department of Engineering, University of Cambridge, 7a JJ Thomson Avenue, Cambridge}
\affil[7]{Research professor, Department of Engineering, University of Cambridge, 7a JJ Thomson Avenue, Cambridge}
\affil[8]{Professor, School of Engineering and Design, Technical University of Munich, Arcisstrasse 21, Munich}
\affil[9]{Professor, Department of Engineering, University of Cambridge, 7a JJ Thomson Avenue, Cambridge}
\maketitle

\begin{abstract}
Ground Penetrating Radar (GPR) is a widely adopted non-destructive sensing technology for subsurface inspection in civil and transportation engineering. Despite its potential for pavement condition assessment, the large-scale application of GPR in automated inspection has two key challenges: the scarcity of annotated real-world datasets and the lack of deep learning models designed for the unique characteristics of 3-Dimensional (3D) GPR data. This study addresses these limitations by firstly introducing a cost-effective data preparation pipeline that integrates orthomosaic Red Green Blue (RGB) imagery with 3D GPR scans to generate annotated 3D GPR datasets. The proposed method uses the aligned segments of RGB and GPR data, using pavement surface images as a reference to transfer labels of surface-visible defects to corresponding GPR segments, enabling efficient large-scale annotation in a real-world dataset collected on a highway section under operation. In addition to the dataset contribution, we propose a specialised 3D Convolutional Neural Network (CNN) architecture incorporating residual connections, mixed convolutional kernel sizes, and both depthwise and channelwise attention mechanisms to enhance feature representation and defect classification. The model is evaluated on binary classification tasks for detecting patch and crack defects in pavement structures. Experimental results demonstrate that the proposed network outperforms baseline architectures across multiple evaluation metrics. Ablation studies further confirm the effectiveness of the designed architectural components. This work contributes a scalable and practical method for real-world dataset generation, along with a novel deep learning framework. It establishes a data foundation for developing more advanced machine learning approaches and benefits the automation and accuracy of GPR-based pavement defect detection in large-scale infrastructure assessment.

\end{abstract}

\section{Practical Applications}
Maintaining roads and pavements is essential for transportation safety and the long-term durability of infrastructure. Ground Penetrating Radar (GPR) is a widely adopted non-destructive sensing technology for subsurface inspection in civil and transportation engineering. Traditionally, identifying subsurface defects such as cracks or voids using GPR data is time-consuming, requires specialised domain expertise, and often involves techniques like drilling. This study presents a new method that combines GPR data with surface images to identify pavement defects. By using high-resolution road imagery to guide the labelling of 3D GPR data, the approach enables the creation of large datasets without manual interpretation of complex GPR scans. A specially designed 3D deep learning model is then trained to recognise defect patterns within this data and outperforms benchmark methods. This paper has three main contributions: a novel data preparation framework that uses an RGB image to annotate GPR data, a 3D real-world annotated dataset, and a novel high-performing deep learning architecture for the data. This brings a significant step toward smarter, data-driven infrastructure management with advanced GPR data analysis.

\section{Introduction}

Road networks are essential infrastructure for economic growth and societal well-being, forming the backbone of the transportation system for moving both goods and passengers.
In 2018, 79\% of domestic freight shipments in the United Kingdom were carried by road \cite{dft2019}.
By 2021, cars, vans, and taxis accounted for $88\%$ of passenger kilometres travelled \cite{dft2021}. 
Highways play a particularly critical role within road networks, carrying substantial traffic volumes. For example, although motorways and "A" roads comprised only $13\%$ of the total road network in the UK by length in 2022, they supported $64\%$ of the total motor vehicle distances travelled \cite{dft2022}.
Given their importance, highways require regular pavement inspections to ensure timely maintenance and reliable infrastructure. 
Without proper upkeep, deteriorating highways can lead to a range of serious consequences, including safety hazards, vehicle damage (e.g., from potholes), and traffic congestion.
These issues not only pose risks to public well-being but also contribute to economic losses and increased pollutant emissions.

In general, there are three categories of pavement assessment: visual inspection, destructive, and non-destructive testing technologies. 
Manual visual inspection, particularly for crack detection in images and videos, is widely adopted in industry, while developing automated vision-based methods is a well-researched area. 
However, such an approach is restricted to detecting visible surface defects, enabling only reactive repairs after apparent deterioration. 
This reactive approach to pavement maintenance can adversely affect highway functionality, as visible damage is often an indicator of advanced deterioration. 
To mitigate these impacts, maintenance practices increasingly require methods capable of detecting potential, yet unobservable defects within pavement structures. Such methods can inform targeted maintenance interventions and reduce traffic disruption. While destructive inspection methods (e.g. pavement corings) are often used to evaluate subsurface pavement conditions, they also present clear disadvantages: they damage the pavement structure, require dedicated operational time windows (often necessitating road closures), and involve relatively high costs and labour demands \cite{RASOL2022126686}.

For these reasons, non-destructive methods, and particularly ground penetrating radar (GPR), are commonly adopted for subsurface inspections of infrastructure \cite{malihi2024}. GPR is a non-destructive technology that is capable of detecting subsurface features, providing insights beyond the visible surface. It has been employed for a wide range of civil engineering tasks, including the detection of hidden utilities and objects \cite{REN2024115379,9540879}, as well as pavement inspection and repair evaluation \cite{LIU2023104689,buildings13071752}.


In road inspection, GPR technology mainly works as a complementary survey combined with more traditional methods, such as visual inspection, drilling, and sampling. 
Several organizations have established standards to guide the application of GPR for road infrastructure inspection.
The American Society for Testing and Materials (ASTM) International published D6432-19 to guide the assessment of subsurface materials using GPR \cite{ASTM}. 
In the United Kingdom, National Highways has published guidance documents on the non-destructive testing of highway structures \cite{DMRB_1} and procedures for pavement assessment \cite{DMRB_2}. 
These guides include detailed requirements and specifications for employing GPR in road and pavement evaluations.


Interpreting GPR data, especially for applications such as detecting pavement structures and assessing subsurface conditions, requires significant expert knowledge and is not an intuitive process. 
The data analysis phase requires substantial manual input, which inhibits scalability for more routine applications.
Additionally, human interpretation introduces variability and the potential for errors, as the analysis is heavily reliant on the expertise of the individual. 
Consequently, there is an increasing demand for automating the interpretation of GPR data to reduce human effort \cite{KUCHIPUDI2022104378}, dependence on expert knowledge, and variability in analysis. 

Deep learning methods have been applied in processing data in road digitalization and condition assessment \cite{lu2025development,LU2026106603}. By leveraging models trained on annotated datasets, GPR data can be analyzed to generate predictions based on patterns and knowledge derived from the training data. 
However, applications of supervised deep learning for interpreting GPR data in large-scale road networks face key challenges, notably the scarcity of annotated datasets.

Synthetic data are used when real data with annotations is expensive to obtain, and have been proven to have their potential to improve the performance \cite{zhou2025impact}. 
Similarly, while synthetic GPR data can be generated using simulation-based approaches, these are limited by their inability to fully capture the complexity and heterogeneity of real-world subsurface conditions, such as variations in material properties and environmental factors \cite{WU2025106275}. As a result, models trained on simulated data often fail to generalize to real-world applications. In addition,  expert 'ground truth' annotation is prone to variability and errors, while physical sampling is often infeasible for roads in active operation. These limitations highlight the need for more scalable and cost-effective approaches to GPR data annotation and interpretation in practical infrastructure applications.

Therefore, the remaining challenges can be summarised from both data and algorithmic perspectives:
\begin{itemize}
    \item Lack of real-world benchmark datasets: Most existing approaches have been tested on datasets collected by the respective authors, making it difficult to evaluate these methods consistently. The absence of a standardized, real-world benchmark dataset hinders the ability to compare different approaches under identical test conditions.
     \item Challenges in data annotation: Annotations are essential for most machine learning-based approaches, yet current methods for annotating GPR data present significant issues. Broadly, there are two primary approaches: (1) annotation through expert interpretation and (2) annotation using test samples from traditional experimental techniques e.g. pavement coring. While expert-based annotation is comparatively faster and cheaper, it depends heavily on the interpreter’s experience, introducing variability and subjectivity. These issues are generally less pronounced in the annotation of other data types, such as images. In contrast, annotations derived from physical testing offer high accuracy and reliability but are labour-intensive, time-consuming, and expensive. Developing efficient and cost-effective annotation strategies is crucial to addressing these limitations and advancing machine learning applications in GPR data analysis.
    \item Limited focus on algorithmic development for real-world 3-Dimensional (3D) GPR data: Despite being more informative, 3D GPR data remains underexplored.  Generic deep learning architectures are not specifically designed for the characteristics of real-world GPR data, where the longitudinal, lateral, and depth dimensions have different physical meanings and noise characteristics. There is a pressing need for deep learning architectures that can directly process real-world 3D GPR data and effectively learn features across these dimensions.
\end{itemize}

To address the abovementioned challenges of annotating GPR data, we propose a novel approach to generating labels by combining pavement images with corresponding GPR data from the same areas. By using visual information from images to guide the annotation of GPR data, we bypass the need for direct annotation of GPR data itself. Image annotation is more straightforward and generates labelled data more efficiently, which is important to apply GPR technology in large-scale road network applications. 
To address the gap in algorithmic development for 3D GPR data, we propose a novel deep learning architecture to improve feature extraction from volumetric GPR data collected from real-world operational roads.

The contributions of this paper can be summarized as follows:

\begin{itemize} 
\item We propose a novel network architecture that integrates 3D convolutional blocks, depth attention mechanisms, and channel attention mechanisms. The proposed architecture achieves robust classification performance on our dataset, demonstrating its effectiveness in processing 3D GPR data. 
\item We introduce a novel dual-modality dataset consisting of co-registered GPR and Red Green Blue (RGB) image samples collected along the longitudinal road direction from a section of concrete pavement on Highway A14 in the UK. 
\item Instead of relying solely on traditional methods such as coring or expert-based ground truth labelling when annotating GPR data, we propose using RGB images as a reference to generate proxy labels for spatially corresponding GPR segments associated with surface-visible pavement defects. Our study shows that deep learning models can learn GPR patterns associated with such defects.

\end{itemize}

The rest of this paper is organised as follows. Research background is provided in Section \ref{sec_background};
The research methodology is introduced in Section \ref{sec_approach};
The experiments and results are presented in Section \ref{sec_result};
conclusions and future work are discussed in Section \ref{sec_conclusion}.

\section{Research Background}
\label{sec_background}

GPR operates by exploiting the varying propagation characteristics of electromagnetic waves as they travel through different media, which are determined by the media's dielectric constants. 
When the signal encounters an interface between two media with significantly different dielectric constants, a portion of the signal is reflected while the remainder continues propagating into deeper layers. These reflected signals are then detected by a receiving antenna to identify subsurface material variations \cite{rs13040672}.
GPR data can be interpreted either by analysing the reflected signals directly or by reconstructing 2-Dimensional (2D) or 3D images of the subsurface. 

\subsection{Signal-based methods}
Traditional methods, compared with recently developed machine learning approaches, can be broadly divided into two categories: signal-based and image-based methods.
In signal-based research, researchers have focused on developing methods to analyze the electromagnetic waves received by GPR antennas. 
Fundamental techniques for processing GPR data, such as filtering, deconvolution, and time-zero correction, are described in \cite{jol2008ground}. 
Building on these foundational methods, numerous advanced techniques have been proposed and applied in pavement assessment.
For instance, regularized deconvolution combined with the L-curve method has been used to determine the thicknesses of various layers in asphalt pavements at a field test site \cite{ZHAO20151}. Similarly, in estimating asphalt density of pavement, a correction algorithm based on baseline GPR scans has been proposed and validated through pavement coring \cite{doi:10.1080/10298436.2014.973027}.

Those methods typically require knowledge and expertise in multiple domains, including signal processing, the operational principles of GPR, and the electromagnetic wave behaviours when propagating through different media. 
Moreover, reliable information extraction often depends on prior knowledge of the scanned targets.
For example, \citeN{rs8050430} employ a Hough transform algorithm for object recognition tasks, which necessitates prior information about the size and orientation of the scanned objects.

\subsection{Image-based methods}
Image-based methods in GPR are primarily employed for processing B-scan and C-scan data.
For detecting subsurface utilities such as pipes and cables, two characteristic patterns—hyperbolic curves and linear segments—are commonly observed in B-scan GPR images \cite{bruschini1998ground}. Numerous methods have been developed to identify these patterns. For instance, \citeN{7572995} proposed a novel clustering algorithm to identify regions of interest, subsequently fitting hyperbolas to these regions. Additionally, the dimensions of buried objects can be derived by analyzing B-scan images; for example, \citeN{CHANG20091057} presented procedures for determining the radius of rebar from B-scan images, validating their approach using several laboratory concrete specimens.

Compared to B-scans, C-scans, representing 3D GPR data generated from a multi-antenna GPR system and composed of multiple B-scans along an orthogonal direction \cite{7055265}, provide a more intuitive and easily interpretable representation \cite{LUO201937}. However, due to the high computational cost associated with processing C-scan data \cite{7055265}, many researchers prefer to simplify the problem into a 2D space. Instead of analyzing hyperboloids in 3D, some studies utilize their 2D projections as hyperbolas and implement separate calculations \cite{1381623,5419982}.
Some studies have also tackled C-scan data directly. For example, \citeN{8090401} propose a feature-based algorithm that highlights changes in reflection values between the layers within C-scan data. 


Similar to signal-based methods, image-based approaches yield reliable results when carefully calibrated, considering factors such as device specifications and the electromagnetic properties of materials. However, a significant challenge remains: these methods heavily rely on expert knowledge. As noted in \citeN{TONG201869}, these techniques are not robust in complex scenarios, often requiring human intervention and assistance.

\subsection{Learning-based approaches}

As a prominent data-driven method, machine learning is increasingly being applied to GPR to automate interpretation. 
Serving as an alternative to traditional signal- and image-based methods, machine learning techniques are generally applied to post-processed GPR data and have been demonstrated to be effective in various civil engineering applications \cite{RASOL2022126686}.

A limited number of studies have explored the analysis of A-scan GPR data using machine learning models. In \citeN{he2017interpretation}, A-scan GPR data collected from tunnels in different regions were transformed into time-frequency maps, which were then classified using a Convolutional Neural Network (CNNs) model. The CNN approach outperformed a Support Vector Machine (SVM) benchmark in terms of accuracy.
Compared to other types of GPR data, interpreting A-scan data independently is less informative and often less effective for detailed analysis.

In contrast, a larger body of research has focused on interpreting B- and C-scans. These data are often treated as raster images, making deep learning approaches, particularly CNNs, an intuitive choice, given the fact that CNNs have proven to be powerful tools for tasks such as image classification and object detection.  

Some researchers use deep learning models to detect underground utilities in B-scan data \cite{SU2023104776}. Moreover, \citeN{TONG2017775} has employed CNNs to identify concealed pavement cracks from B-scan GPR data, using a dataset labeled with information derived from pavement core sampling. In subsequent studies, additional CNN-based variations \cite{TONG201869} and network-in-network architectures \cite{TONG2020117352} are applied to datasets annotated with additional classes such as subgrade unevenness and subgrade sinkholes. Several popular CNN architectures have also been adopted in this field. For instance, the AlexNet architecture \cite{krizhevsky2012imagenet} has been applied to pre-processed B-scan data for the detection of underground objects in urban areas \cite{doi:10.1080/10298436.2018.1559317}. Various You Only Look Once (YOLO)-based frameworks \cite{redmon2016you} have also been employed to detect underground utilities such as cables and pipes \cite{isprs-archives-XLII-2-W16-293-2019}, as well as concealed cracks and moisture damage in asphalt pavements \cite{LI2021121949,ZHANG2020103119}. Furthermore, faster Region-based CNN frameworks have been explored for identifying buried objects by detecting hyperbolic reflections \cite{8517683} and pavement distress such as cracks, water-damaged pits, and uneven settlements \cite{GAO2020108077}.
 

Beyond CNNs, other architectures such as recurrent neural networks (RNNs) have been adapted for B-scan data processing. \citeN{electronics9111804} integrated CNNs with Long Short-Term Memory (LSTM) networks, a type of RNN, to detect hyperbolic region features and estimate the diameters of buried cylindrical objects. In \cite{10546904}, a network based on the vision transformer architecture is proposed to classify pavement distress severity.


Although C-scan data provide more comprehensive and informative representations, fewer studies have focused on C-scan data. In \citeN{ZHOU2024105819}, the authors propose a network architecture integrating U-Net, residual networks, and attention mechanisms to segment 3D C-scan data and reconstruct underground pipes. Other studies \cite{doi:10.1080/10298436.2019.1645846,rs11212545} utilize collections of B-scan and C-scan images saved as 2D grid images. These are then classified using CNN-based networks into categories such as cavities, pipes, and manholes. Similarly, in \citeN{LIU2023104698}, one B-scan and one C-scan image are stitched together and input into a YOLO-based network to identify cracks in pavements.

\subsection{Research gaps}
The above review shows that existing GPR interpretation methods have progressed from traditional signal- and image-based analysis to learning-based approaches. However, several challenges remain. Firstly, learning-based approaches mostly require annotated datasets. Currently, real-world annotated GPR datasets for pavement assessment are still limited, and the annotation process relies on expert interpretation or physical validation, both of which are difficult to scale across operational road networks. Secondly, from an algorithmic perspective, many learning-based methods focus on A-scan or B-scan data, or simplify 3D C-scan data into 2D representations. Such simplification may limit the ability of models to learn 3D volumetric information from the data.

To address the data gap, this paper proposes an RGB-guided annotation pipeline that transfers labels from pavement images to spatially corresponding 3D GPR segments. To address the algorithmic gap, this paper develops a specialised 3D convolutional neural network incorporating residual connections, mixed convolutional kernels, and depth-wise and channel-wise attention mechanisms for learning discriminative features from real-world 3D GPR data.

\section{Methodology}
\label{sec_approach}

This section outlines the proposed pipeline, which includes generating a 3D GPR dataset through RGB-guided annotation and designing a deep learning architecture to interpret the annotated GPR data.

\subsection{Dataset preparation}
\label{subsec:dataset}
\begin{figure}[H]
         \begin{subfigure}[b]{0.5\textwidth}
	\centering
  	\includegraphics[width=\textwidth]{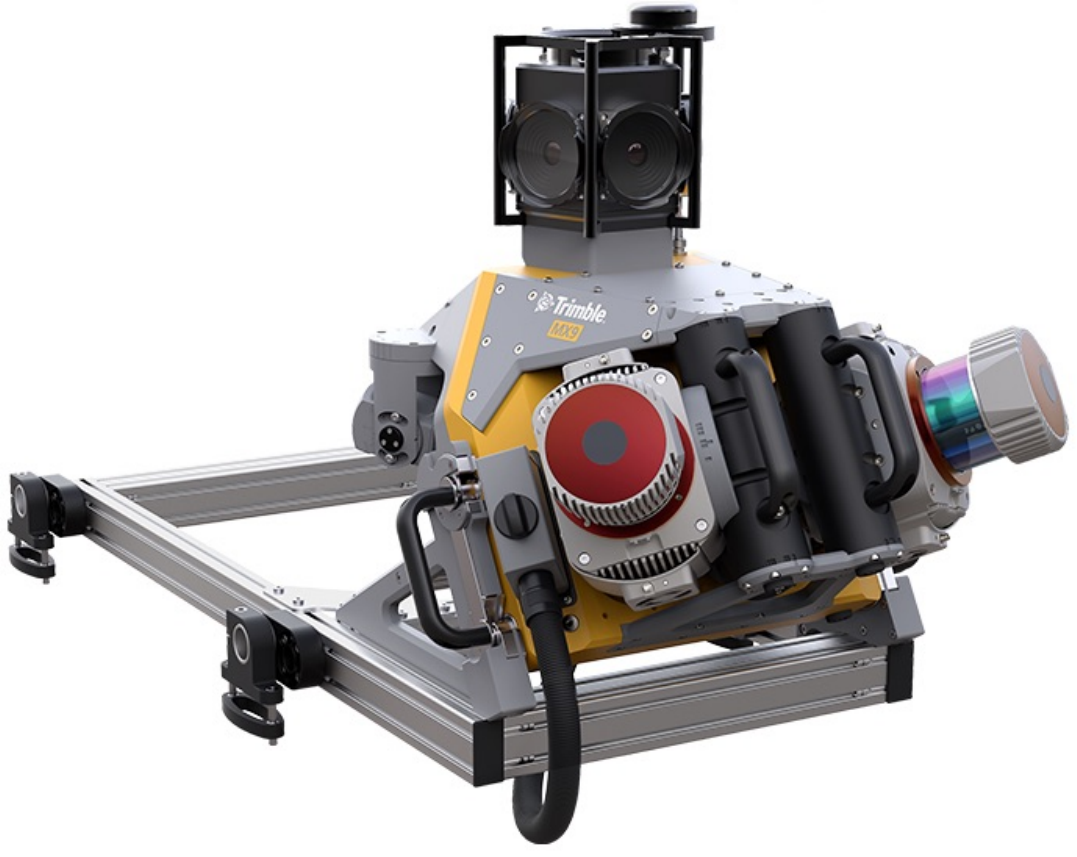}
  	\caption{Trimble MX9 mobile mapping system (image source Trimble \cite{mx9})}
  	\label{fig:mx9}
        \end{subfigure}
        \begin{subfigure}[b]{0.5\textwidth}
	\centering
  	\includegraphics[width=\textwidth]{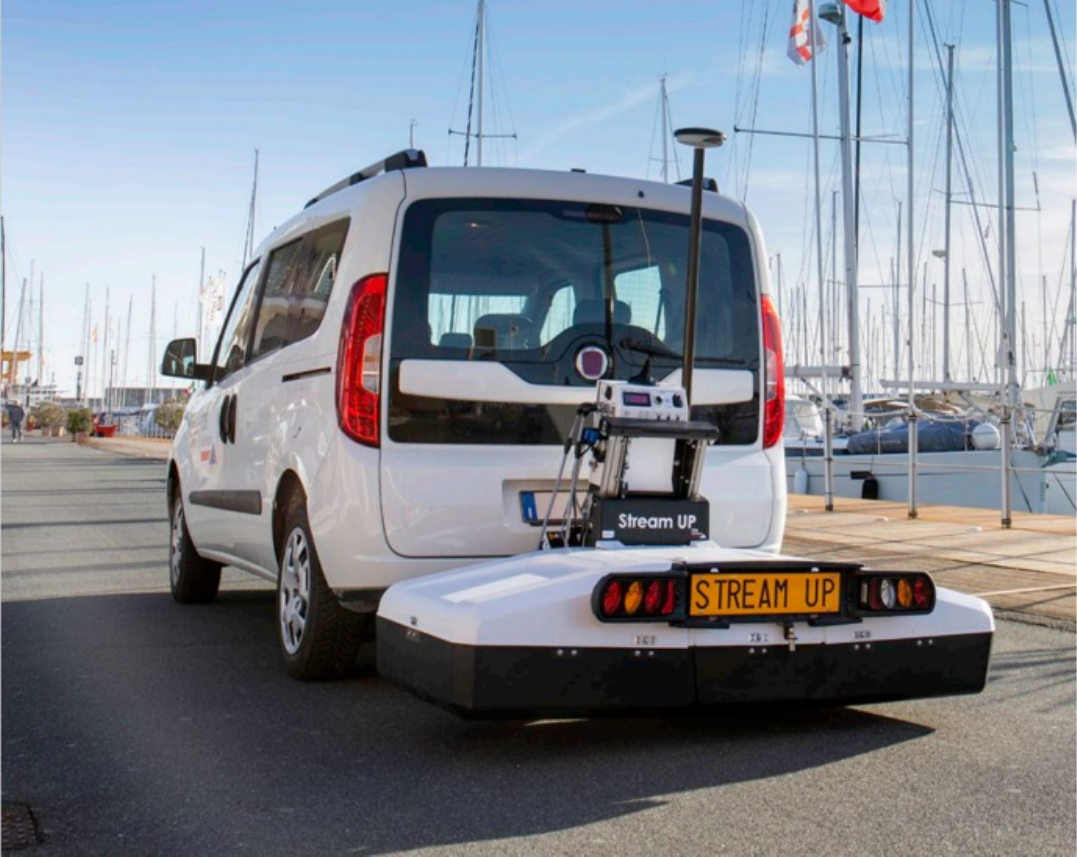}
  	\caption{Hexagon Stream Up GPR system (Image source: Hexagon \cite{streamup})}
  	\label{fig:streamup}
        \end{subfigure}   
  	\caption{Equipment used in CAMHighways Dataset  \protect\cite{DRFdataset}}
  	\label{fig:equipment}
\end{figure}

\begin{figure}[H]         
            \centering
            \includegraphics[width=\linewidth]{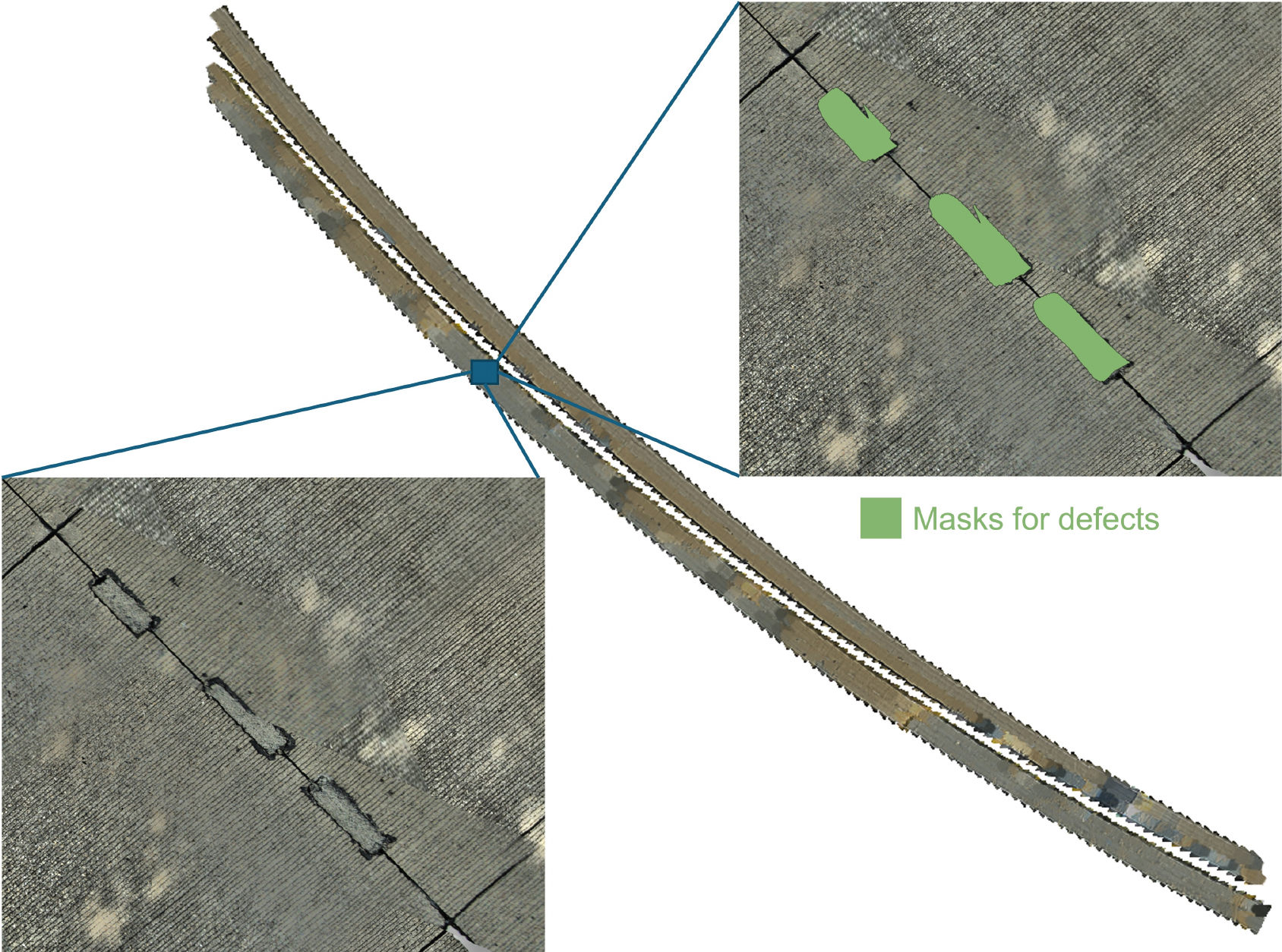}
            \caption{Orthomosaics built from the pavement images and defect masks}
            \label{fig:rawortho}  
\end{figure}

The GPR data used in this study is part of the CAMHighways Dataset \cite{DRFdataset}, collected from various sections of the UK highway network. This dataset includes point cloud data, RGB images, thermal images, and GPR data.
The mobile mapping data were acquired in 2021 using a Trimble MX9 mobile mapping system, while the GPR data were collected later with a Hexagon Stream Up GPR system. Both data acquisition systems are illustrated in Figure \ref{fig:equipment}.
Orthomosaics, showing a top-down view of pavements generated from the pavement images after removing camera corrections, defect mask orthomosaics generated from annotations of pavement images, and GPR data are all georeferenced to the British National Grid (EPSG:27700 or OSGB36).
Figure \ref{fig:rawortho} shows example orthomosaics alongside their corresponding defect masks from the CAMHighways dataset. 

\begin{figure}[H]
         \begin{subfigure}[b]{0.4\textwidth}
	\centering
  	\includegraphics[width=\textwidth]{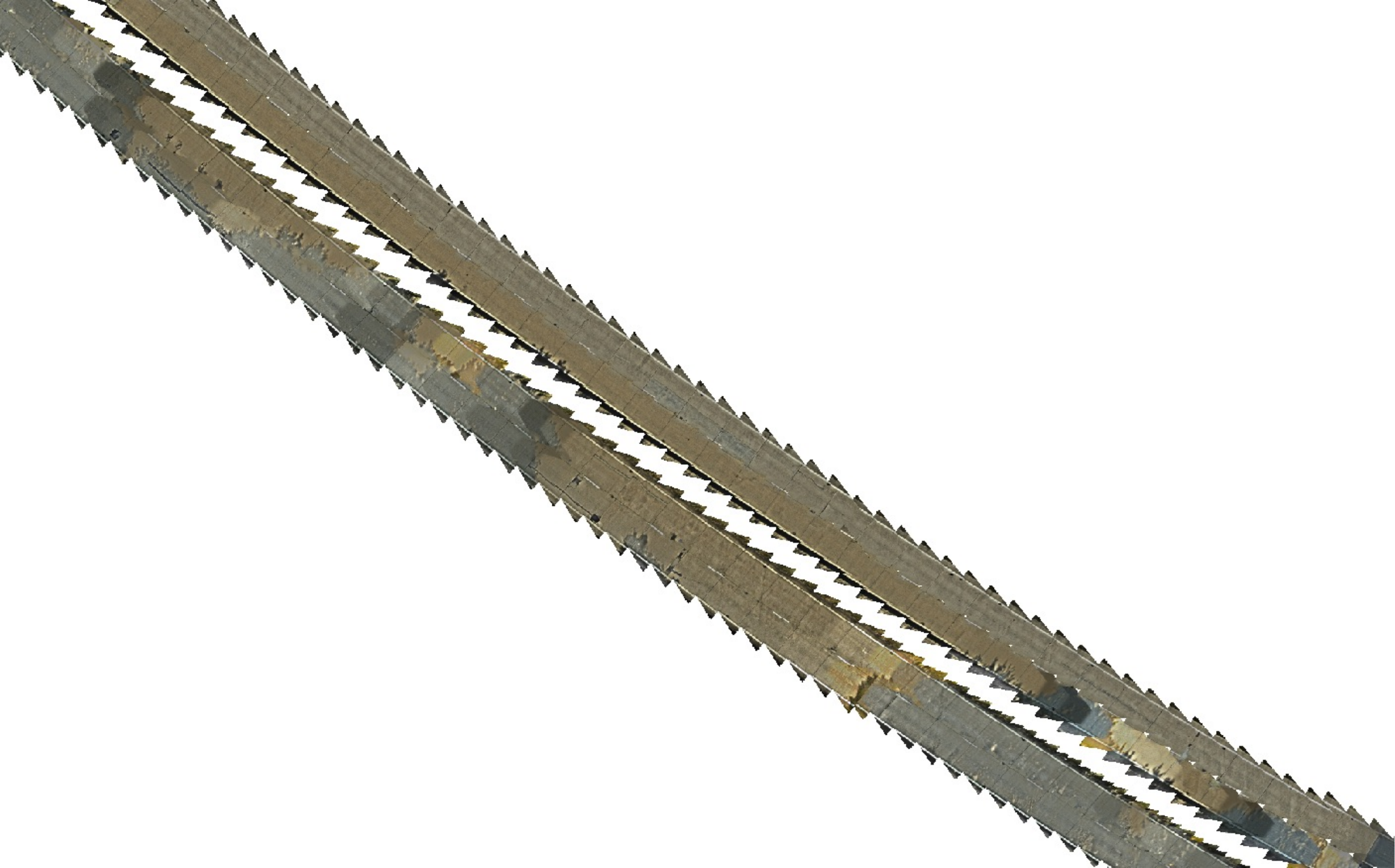}
  	\caption{Exemplar orthomosaics of one road section}
  	\label{fig:ortho_gpr_1}
        \end{subfigure}
       \begin{subfigure}[b]{0.4\textwidth}
	\centering
  	\includegraphics[width=\textwidth]{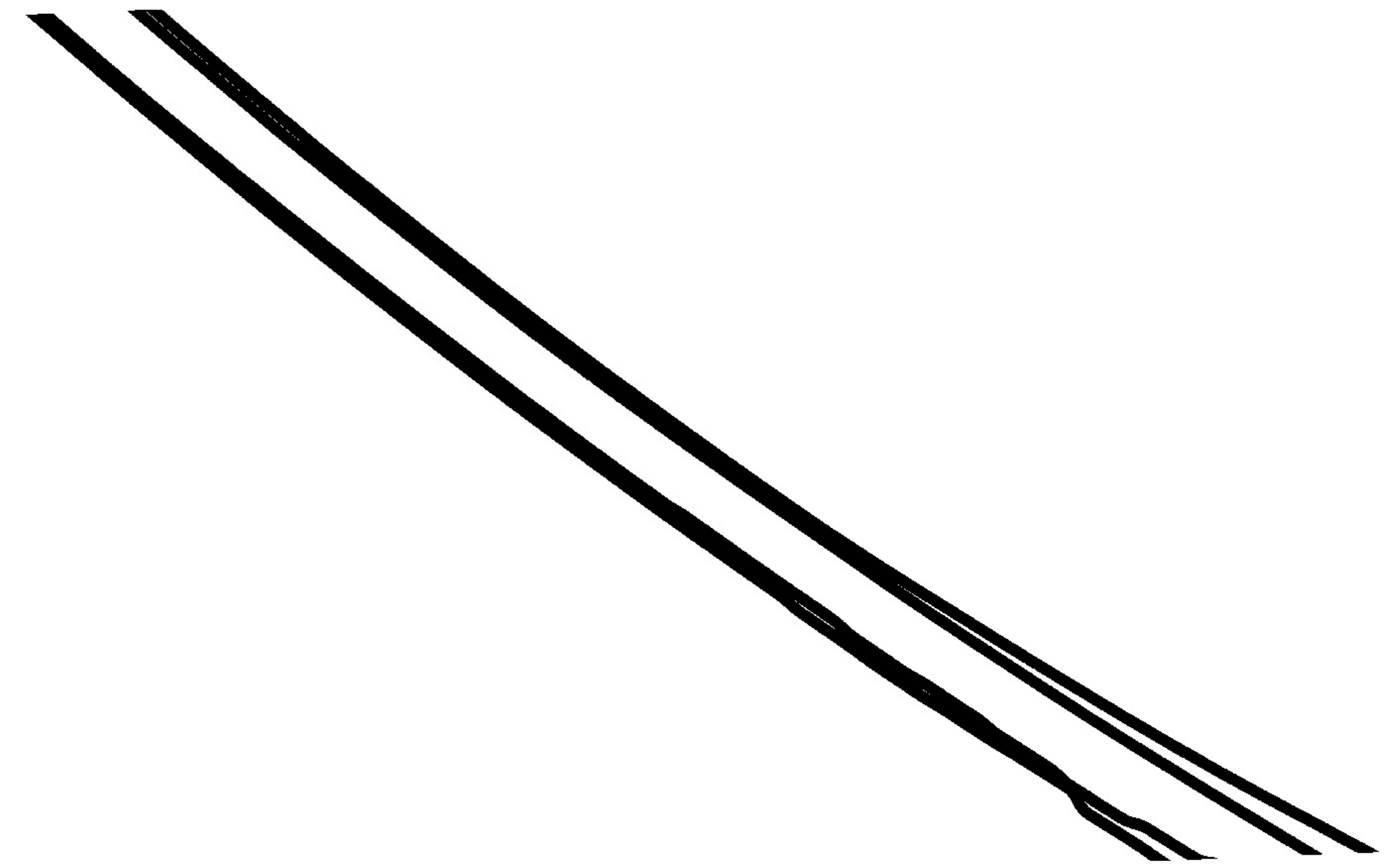}
  	\caption{Exemplar GPR data coverage for the same road section}
  	\label{fig:ortho_gpr_2}
        \end{subfigure}
        \begin{subfigure}[b]{0.4\textwidth}
	\centering
  	\includegraphics[width=\textwidth]{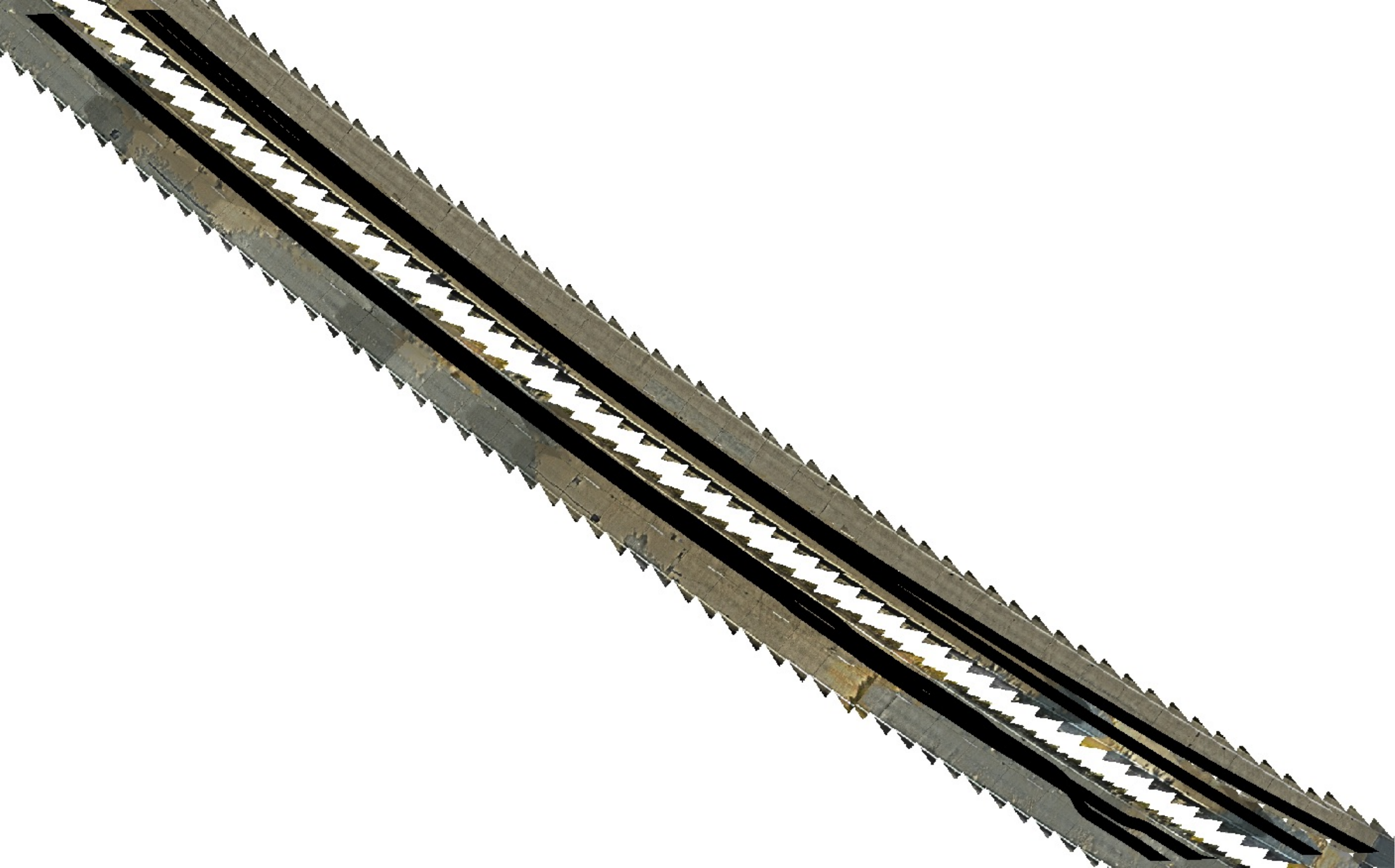}
  	\caption{Overlapped orthomosaics and GPR data}
  	\label{fig:ortho_gpr_3}
        \end{subfigure}
  	\caption{Illustration of spatial misalignment between orthomosaics and GPR data for the same road section}
  	\label{fig:orthogpr}
\end{figure}

The present GPR dataset was acquired by the above-mentioned GPR system at a different time from the image data collected by the mobile scanning system, resulting in differences in the routes driven by the survey vehicle. The pavement images were captured lane by lane, with a single drive covering the imagery for one lane. However, the GPR system, with a coverage width of 1.58 m \cite{streamup}, could not cover an entire lane in a single pass, requiring multiple drives to scan one lane fully. The coverage for both data sources is illustrated in Figure \ref{fig:orthogpr}. While Figure \ref{fig:ortho_gpr_1} shows an example orthomosaic from the dataset, Figure \ref{fig:ortho_gpr_2} depicts the corresponding GPR data coverage for the same area. Notably, Figure \ref{fig:ortho_gpr_3} shows the overlap between the GPR coverage and orthomosaics. 

\begin{figure}[H]
         \begin{subfigure}[b]{\textwidth}
	\centering
  	\includegraphics[width=\textwidth]{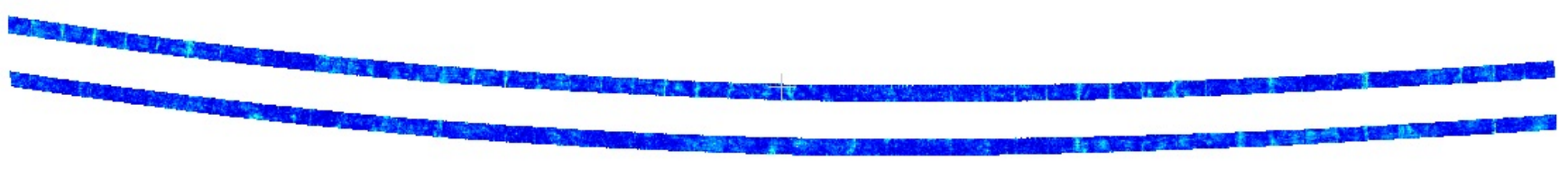}
  	\caption{2D GPR data at the road surface (z = 0 cm)}
  	\label{fig:gpr_1}
        \end{subfigure}
       \begin{subfigure}[b]{\textwidth}
	\centering
  	\includegraphics[width=\textwidth]{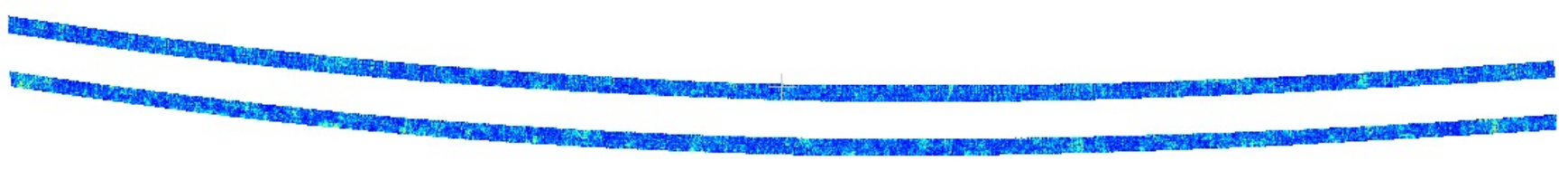}
  	\caption{2D GPR data underground at a depth of z = -95 cm}
  	\label{fig:gpr_2}
        \end{subfigure}
        \begin{subfigure}[b]{\textwidth}
	\centering
  	\includegraphics[width=\textwidth]{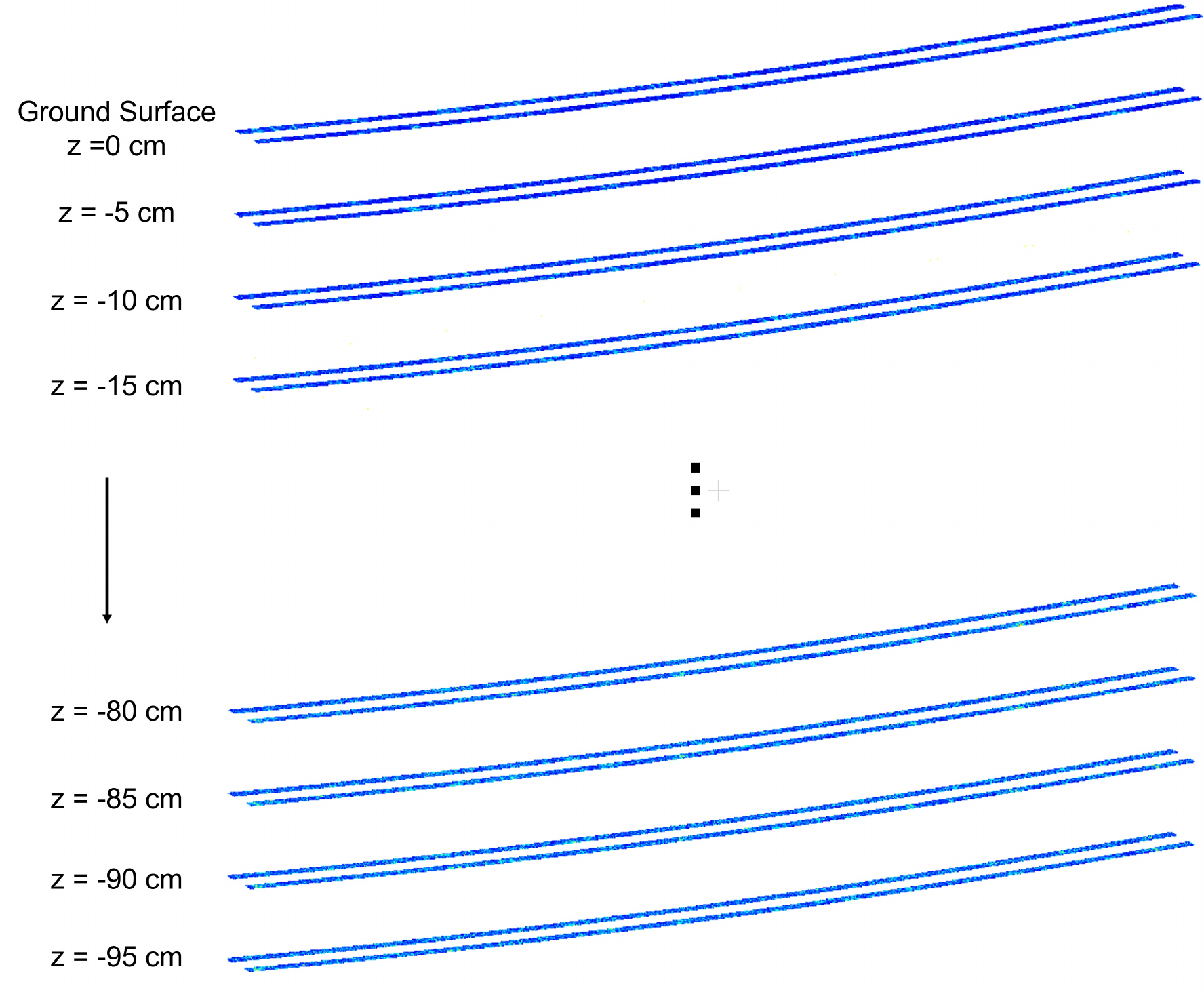}
  	\caption{20 layers of GPR data structured in 3D}
  	\label{fig:gpr_3}
        \end{subfigure}
  	\caption{Exemplar raw GPR data collected from two lanes of road section}
  	\label{fig:gprraw}
\end{figure}

In this paper, a section of concrete pavement on A14 in the UK, from Woolpit to Haugley, is used to build the image and GPR dataset.
The collected GPR data is organized into 20 layers depthwise: from the ground surface to a depth of 95 cm below ground level (in increments of 5 cm). Exemplar GPR data collected across two lanes of a road section is shown in Figure \ref{fig:gprraw}. Figure \ref{fig:gpr_1} presents the 2D GPR data corresponding to the road surface (z = 0 cm), while Figure \ref{fig:gpr_2} illustrates the 2D data at a depth of z = -95 cm. Figure \ref{fig:gpr_3} depicts how these layers are structured in 3D, with some intermediate layers omitted for clarity.

It should also be noted that the GPR data used in this study are processed C-scan volumes exported from the manufacturer's acquisition and processing pipeline, rather than raw time-domain A-scan traces. The manufacturer's processing pipeline is understood to perform standard GPR preprocessing operations, such as time-zero correction, de-wow filtering, background or direct-wave removal, gain compensation, and band-pass filtering, before generating the C-scan slices. The exact numerical parameters of these proprietary preprocessing steps were not available to the authors. 


In addition, the vertical coordinate of raw GPR data is originally represented by two-way travel time rather than physical depth. In the dataset used in this study, the time-to-depth conversion was performed internally by the manufacturer's proprietary processing pipeline, rather than through field calibration by the authors. The specific effective dielectric constant used in this conversion was not available to the authors and is not reported in the publicly available CAMHighways documentation. Because the survey was conducted on a section of highway under continuous operation, on-site coring for dedicated wave-velocity calibration was not feasible. Therefore, the reported depths should be interpreted as nominal depth values, and their absolute accuracy may vary depending on local variations in moisture content and material composition. This does not affect the main methodological contribution of this study, because the proposed framework focuses on RGB-guided annotation and learning from co-registered GPR segments rather than estimating the absolute depth of subsurface objects.

\begin{figure}[H]         
            \centering
            \includegraphics[width=\linewidth]{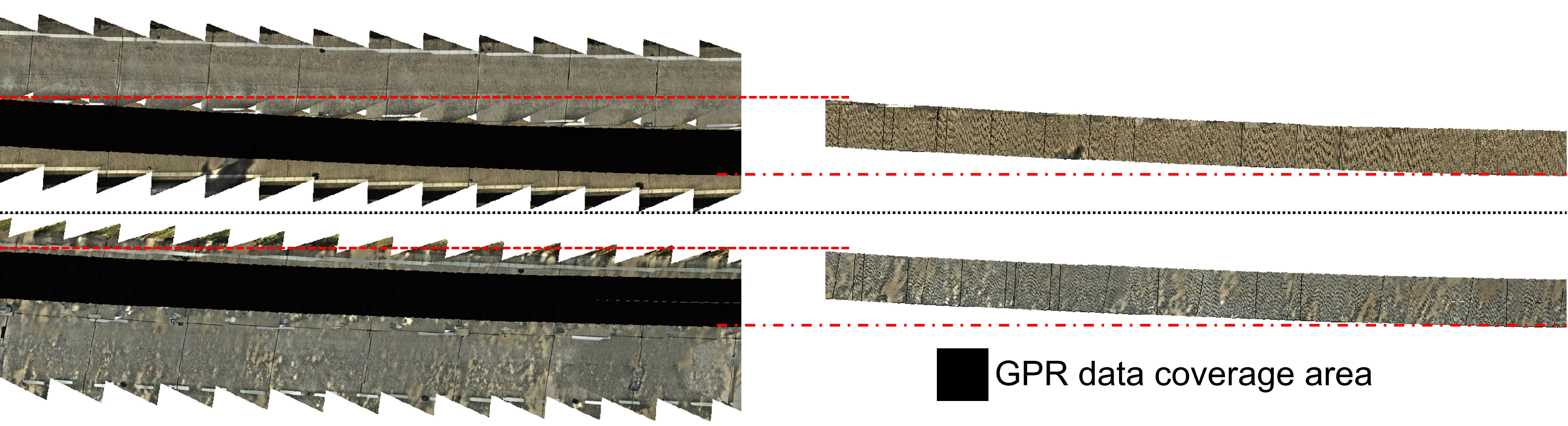}
            \caption{Extraction of overlapping area from example orthomosaics and GPR data}
            \label{fig:overlap}  
\end{figure}

\begin{figure}[H]         
            \centering
            \includegraphics[width=\linewidth]{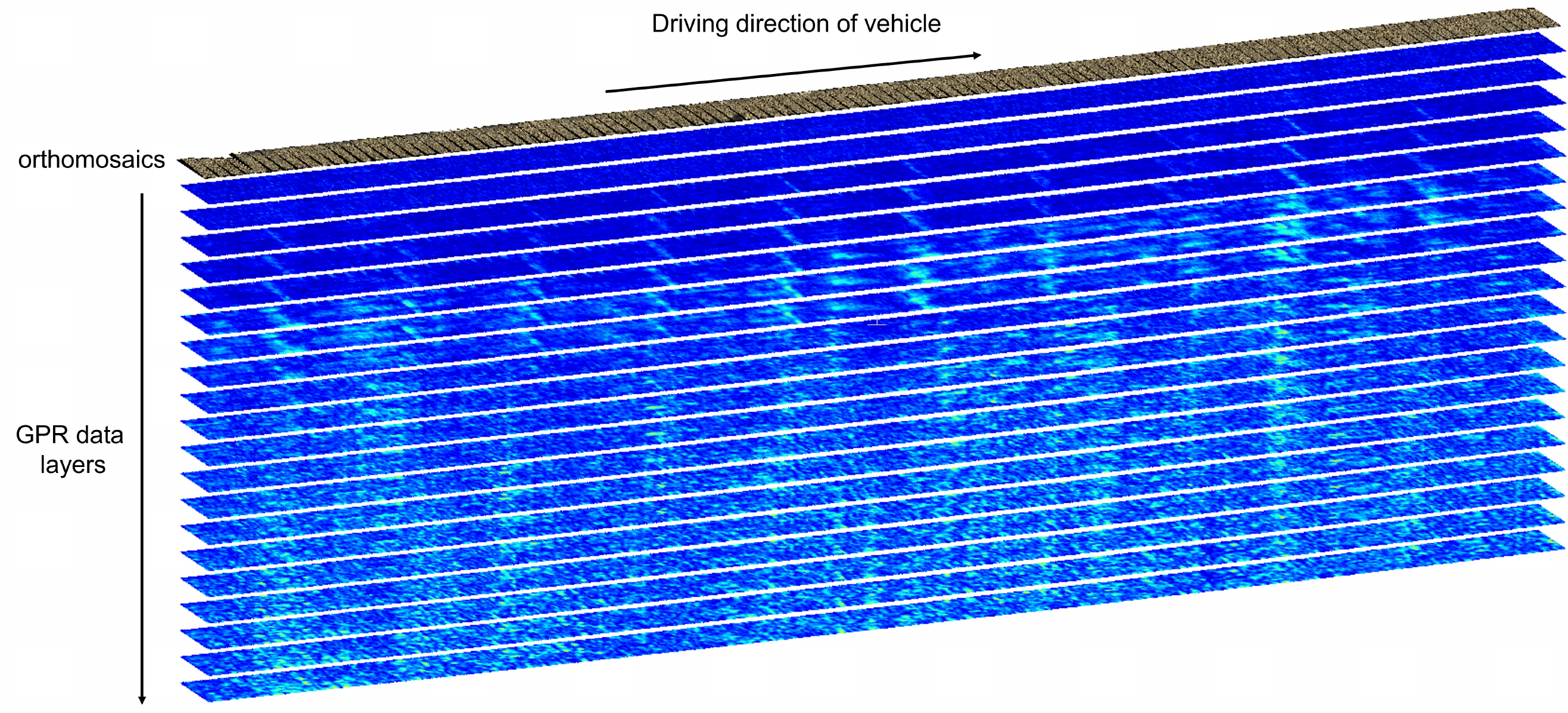}
            \caption{Extracted orthomosaics and GPR data layers for one lane}
            \label{fig:lanedata}  
\end{figure}

The dataset preparation involves computing the overlapping areas between the orthomosaics and GPR data, followed by separating the areas into individual lane clusters based on their Euclidean distance. It should be noted that this method relies on the assumption that GPR data corresponding to individual lanes are always spatially separated, which is the case in the CamHighway dataset. Figure \ref{fig:overlap} illustrates example orthomosaics and GPR data collected over two lanes, along with the extracted overlapping area. Given that both RGB pavement images and GPR data were collected and organized as the survey vehicle passed the road lane by lane, the RGB-GPR dataset was constructed on a per-lane basis. This approach treats the data for each lane, along the direction of the roadway, as a separate and independent unit for subsequent processing. Figure \ref{fig:lanedata} shows an example of the extracted orthomosaics and GPR data layers for a single lane.

\begin{figure}[H]         
            \centering
            \includegraphics[width=0.6\linewidth]{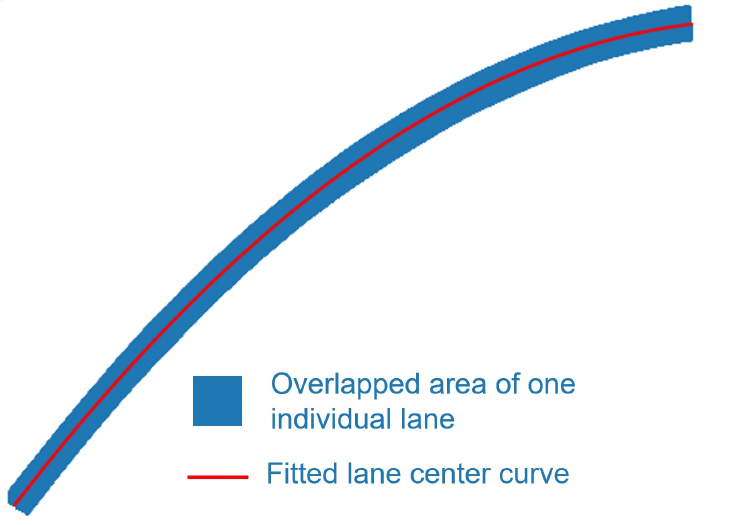}
            \caption{Overlapping area of orthomosaics and GPR coverage and corresponding fitted centerline}
            \label{fig:fit}  
\end{figure}

For each lane, the next step involves generating data segments by cutting the lane’s data orthogonal to its centerline direction. This process begins by fitting a centerline curve along the lane’s longitudinal direction. Since the lane is not always perfectly straight, the centerline is modeled using a third-order polynomial, expressed as:
\begin{equation}
    y = \sum_{k=0}^{3} p_k x^k
\end{equation}
where \( x \) and \( y \) are the coordinates of points within the overlapping area of the orthomosaic and GPR coverage on the 2D ground plane, and \( p_k \) are the coefficients of the polynomial. An example of a lane’s overlapping area and the corresponding fitted centerline is shown in Figure \ref{fig:fit}.  

Subsequently, key points are selected along the fitted centre line at a predefined interval, denoted as $\delta$, representing the length of each segment along the lane direction. In this study, $\delta$ is set to 0.3 meters, which offers a practical balance between spatial resolution and computational efficiency. 
Each key point is denoted as $P_i=(x_i,y_i)$, representing the coordinates of the $i^{th}$ point along the fitted centreline. The corresponding components of the normal vector to the point of the segment are denoted $n_i=(n_{x_i},n_{y_i})$.
These key points define the centres of the final data segments along the lane direction.
All points within the overlapped area are assigned to segments based on their nearest normal projection. 
The distance from a point $Q_p=(x_p,y_p)$ in the overlapped area to the $i^{th}$ segment can be calculated as:

\begin{equation}
    d_{p,i}= \mid (x_p-x_i) \cdot n_{y_i} - (y_p-y_i) \cdot n_{x_i}\mid.
\end{equation}

Each point $Q_p$ is assigned to the segment whose centreline point is closest to it, based on the minimum distance $d_{p,i}$. The index of the nearest segment is determined as
\begin{equation}
    i^* = \underset{i}{\operatorname{argmin}} \, d_{p,i}.
\end{equation}

\begin{figure}[H]         
            \centering
            \includegraphics[width=0.8\linewidth]{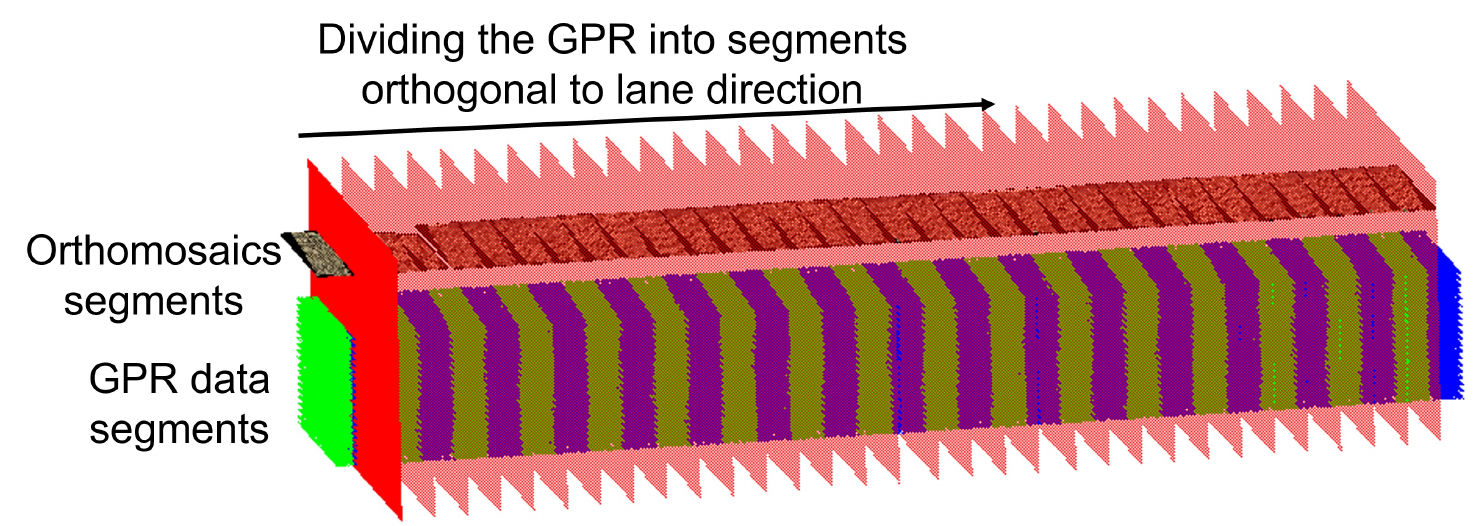}
            \caption{Composite dataset of one lane divided into segments using one moving plane orthogonal to lane direction}
            \label{fig:chopdata}
\end{figure}

Through this process, each point $Q_p$ in the overlapped area is assigned to a segment. The final dataset can thus be viewed as the result of dividing the lane into a series of segments using a moving plane orthogonal to the lane’s centreline along its length, which is illustrated in Figure \ref{fig:chopdata}, where different colours distinguish adjacent segments.
The corresponding pseudocode is provided in Algorithm \ref{algo1}. Each segment has a size of \(100 \times 25 \times 20 \times 3\), where 100 represents the length along the lane (longitudinal direction), 25 represents the lateral direction across the lane, 20 is the vertical depth, and 3 denotes the RGB color channels.

\begin{algorithm}[H]
\caption{Assign GPR Points to Nearest Segment}
\label{algo1}
\begin{algorithmic}
\State \textbf{Inputs:}
\State \hspace{1em} Key Points: List of key points along the fitted central curve, \( P_1, \ldots, P_n \), where each \( P_i = (x_i, y_i) \) 
\State \hspace{1em} Normal Vectors: List of normal vectors at each key point,  \( \mathbf{n}_1, \ldots, \mathbf{n}_n \), where each \( \mathbf{n}_i = (n_{x_i}, n_{y_i}) \) is the normal vector at \( P_i \) 
\State \hspace{1em} GPR Points: List of GPR points to be assigned, \( Q_1, \ldots, Q_m \), where each \( Q_p = (x_p, y_p) \) is a point to be assigned 
\State \textbf{Outputs:}
\State \hspace{1em} Segment Indices: List of indices \( i^* \) representing the closest segment for each GPR point
\State 
\State \textbf{Initialization:}
\State  Create an empty list \texttt{segment\_indices} to store the index \( i^* \) of the closest segment for each GPR point.
\State 
\For{each GPR Point \( Q_p = (x_p, y_p) \) in the list of GPR Points}
    \State Set \texttt{min\_distance\_Q\_p} to a very large value (\(\infty\)).
    \State Set \texttt{best\_segment\_index\_Q\_p} to \(-1\).
    \For{each Segment \( i \) (Defined by \( P_i = (x_i, y_i) \) and \( n_i = (n_{x_i}, n_{y_i}) \))}
        \State Compute the distance \( d_{p,i} \) from \( Q_p \) to segment \( i \) using the formula:
        \[
        d_{p,i} = \left| (x_p - x_i) \cdot n_{y_i} - (y_p - y_i) \cdot n_{x_i} \right|
        \]
        \If{ \( d_{p,i} \) is less than \texttt{min\_distance} }
            \State Set \texttt{min\_distance\_Q\_p} to \( d_{p,i} \).
            \State Set \texttt{best\_segment\_index\_Q\_p} to \( i \).
        \EndIf
    \EndFor
    \State Append \texttt{best\_segment\_index\_Q\_p} to \texttt{segment\_indices}.
\EndFor
\State 
\State \textbf{Output:}
\State \hspace{1em} The list \texttt{segment\_indices} contains the index \( i^* \) of the nearest segment for each GPR point \( Q_p \).
\end{algorithmic}
\end{algorithm}

\begin{figure}[H]
         \begin{subfigure}{0.3\textwidth}
	\centering
  	\includegraphics[width=\linewidth]{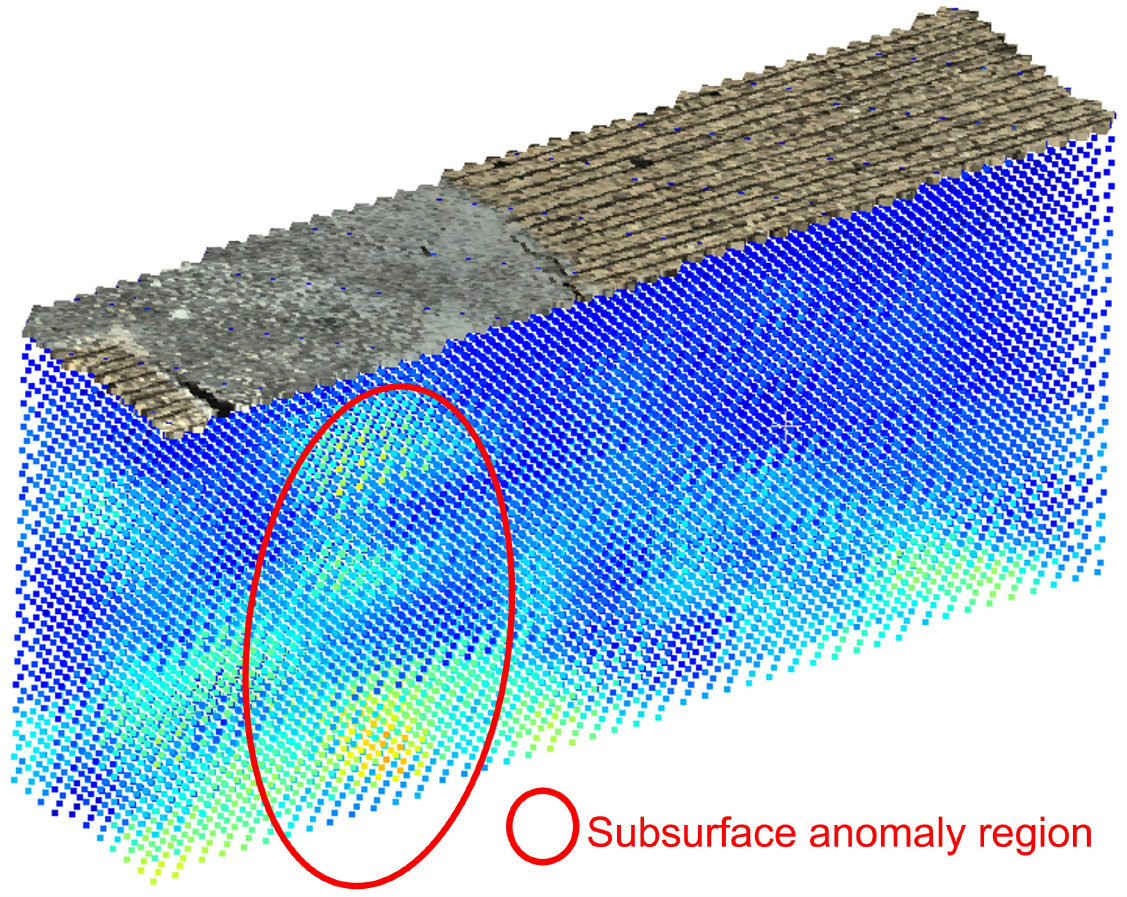}
  	\caption{Patch segment with anomaly marked}
  	\label{fig:patchslice}
        \end{subfigure}      
        \begin{subfigure}{0.3\textwidth}
	\centering
  	\includegraphics[width=\linewidth]{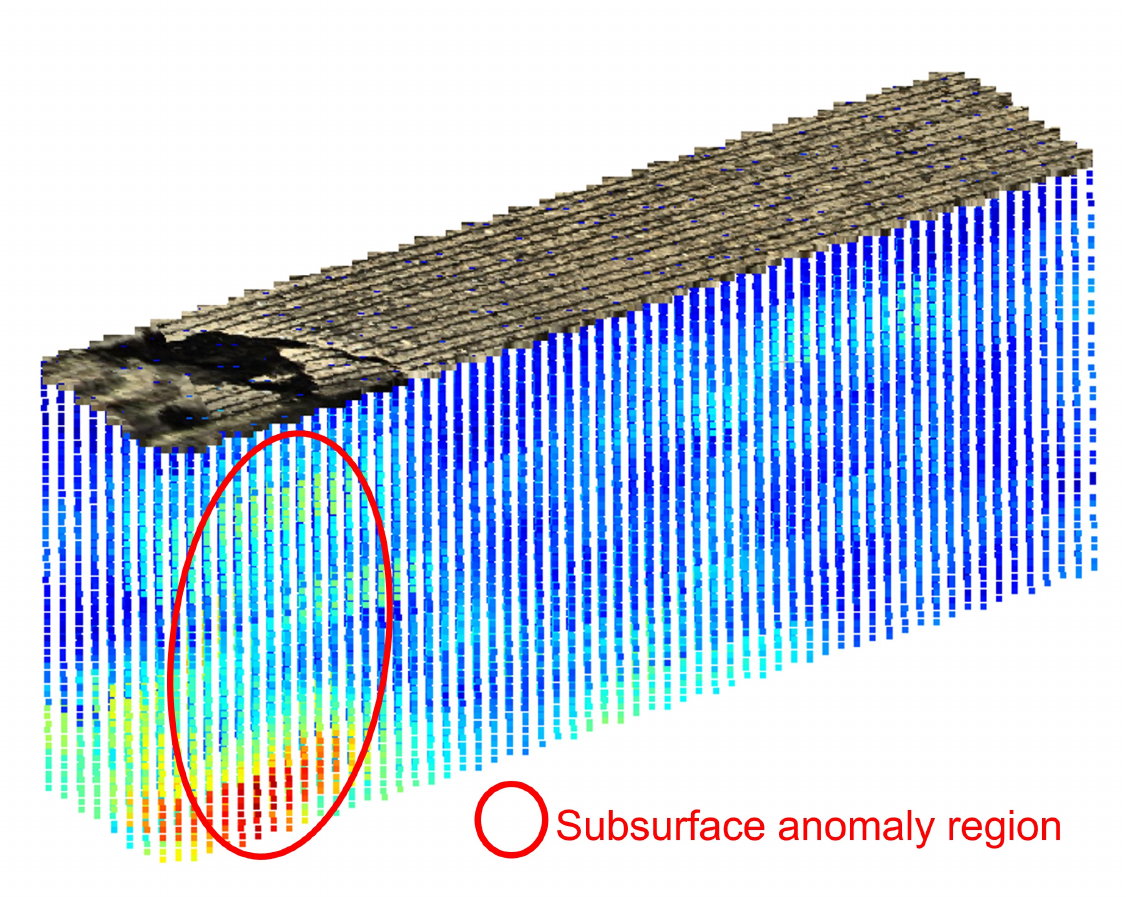}
  	\caption{Crack segment with anomaly marked}
  	\label{fig:crackslice}
        \end{subfigure}        
        \begin{subfigure}{0.3\textwidth}
            \centering
            \includegraphics[width=\linewidth]{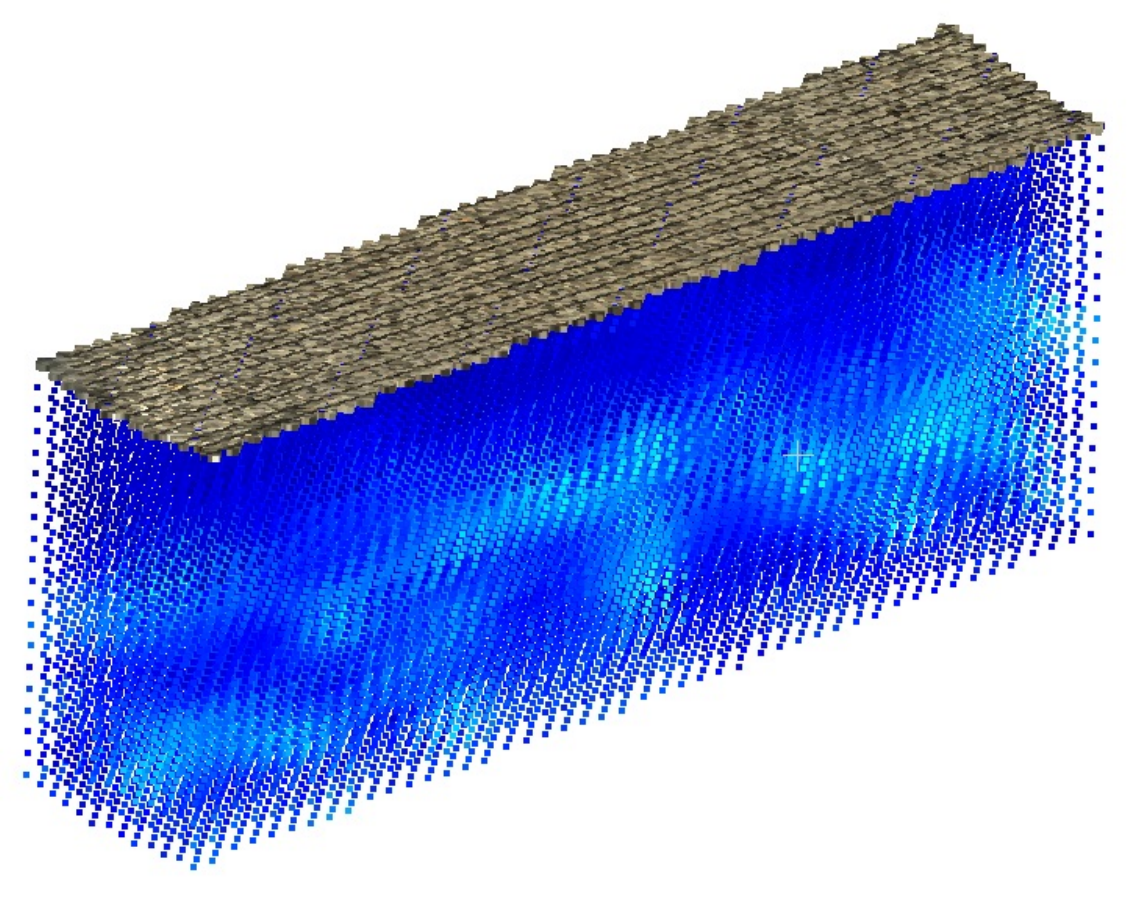}
            \caption{Non-defect segment}
            \label{fig:nondefect}
        \end{subfigure}
        \begin{subfigure}{\textwidth}
	\centering
            \includegraphics[width=\linewidth]{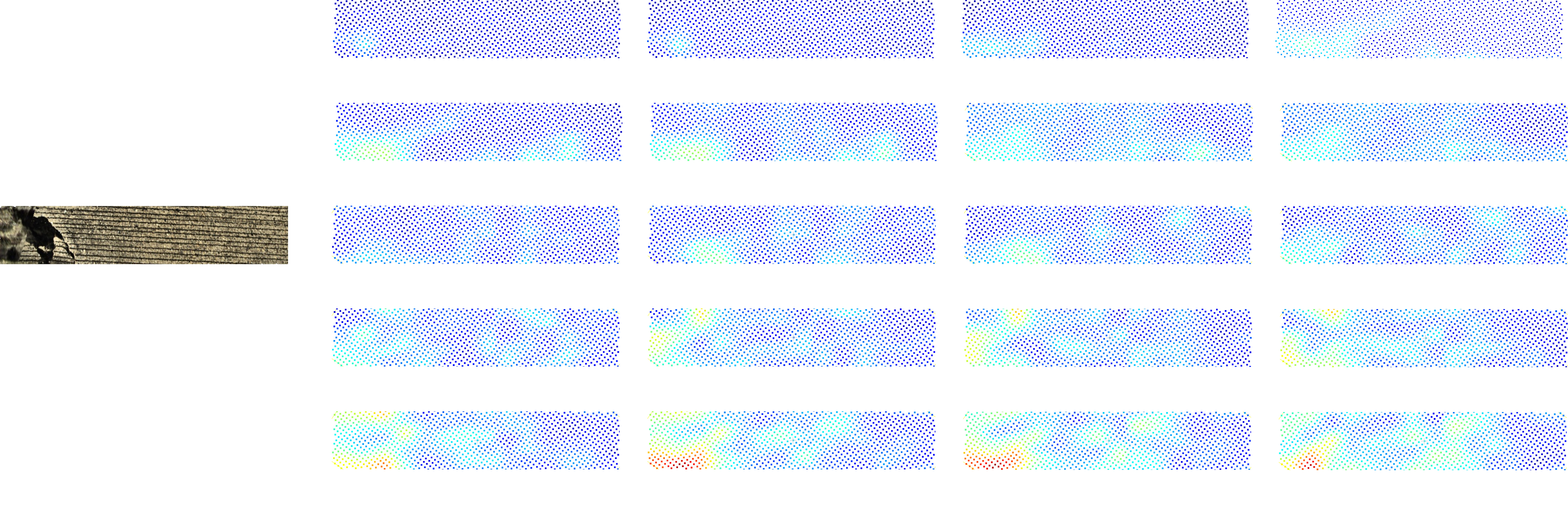}
            \caption{Pavement image and GPR layers in 2D}
            \label{fig:gprlayer}  
        \end{subfigure}  
        \caption{Visualisation of prepared data segments}
        \label{fig:datasegments}  
\end{figure}

Figure \ref{fig:datasegments} presents visualisations of example data segments in the final constructed dataset. Figure \ref{fig:patchslice}, \ref{fig:crackslice}, and \ref{fig:nondefect} show segments containing a patch defect, a crack defect, and no defect, respectively, as seen on the orthomosaics. Figure \ref{fig:gprlayer} illustrates the orthomosaic of a segment alongside its corresponding GPR data layers, displayed layer by layer.

\subsection{Network design}
\label{subsec:network}

\begin{figure}[H]         
            \centering
            \includegraphics[width=\linewidth]{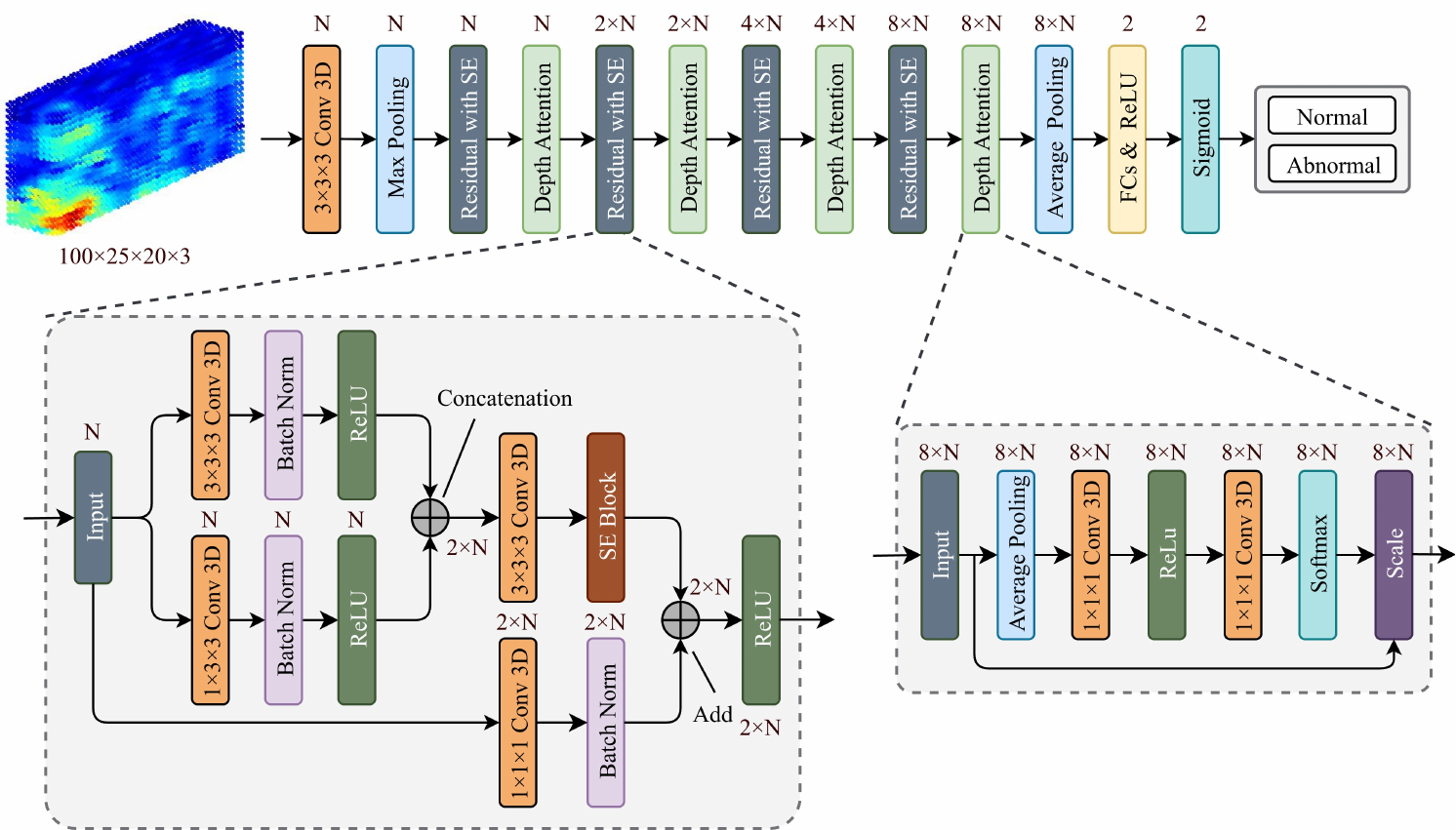}
            \caption{Proposed network architecture for GPR segment classification tasks}
            \label{fig:gprnetwork}  
\end{figure}

A 3D CNN with residual connections and attention mechanisms is designed for the classification of 3D GPR segments. 
This network uses a combination of advanced building blocks, including mixed-kernel residual modules, and depthwise feature attention mechanisms, to enhance feature representation and learning efficiency. 
The designed architecture is shown in Figure \ref{fig:gprnetwork}.

\subsubsection{Residual block with mixed kernel convolutions and channel-wise feature attention}

This module is a custom 3D residual block designed to enhance feature representation by combining mixed kernel sizes of convolutional computing and Squeeze-and-excitation (SE) attention \cite{Hu_2018_CVPR}.
It has two parallel convolutional paths to extract features with different receptive fields and leverages an SE block to recalibrate channel-wise feature responses. 
A residual connection is employed to preserve input information and facilitate gradient flow during training.

The input tensor to the block is denoted \( X \in \mathbb{R}^{B \times C_{\text{in}} \times D \times H \times W} \), where \( B \), \( C_{\text{in}} \), \( D \), \( H \), and \( W \) represent the batch size, number of input channels, depth, height, and width, respectively. 
The block processes the input using two parallel paths. 

The first path applies a \( 3 \times 3 \times 3 \) convolution with kernel \( W_1 \in \mathbb{R}^{(C_{\text{out}} / 2) \times C_{\text{in}} \times 3 \times 3 \times 3} \), 
with stride \( s \), and padding \( p = 1 \), 
followed by a batch normalization layer \(\text{BN}_1(\cdot)\) and a ReLU activation function:
\begin{equation}
\text{Path}_1(X) = \text{ReLU}\big(\text{BN}_1(W_1 \ast X + b_1)\big),
\end{equation}
where \( b_1 \in \mathbb{R}^{C_{\text{out}} / 2} \) is the bias term, \(C_{\text{out}} / 2 \) denotes the number of output channels (or filters) produced by the convolution layer, and \( \ast \) denotes the convolution operation. 
This produces an output tensor of dimensions \( \text{Path}_1(X) \in \mathbb{R}^{B \times (C_{\text{out}} / 2) \times D' \times H' \times W'} \), where \( D', H', W' \) are the spatial dimensions of the output result. 

The second path applies a mixed kernel convolution using a \( 1 \times 3 \times 3 \) kernel. The operation uses kernel \( W_2 \in \mathbb{R}^{(C_{\text{out}} / 2) \times C_{\text{in}} \times 1 \times 3 \times 3} \), 
followed by a batch normalization layer \(\text{BN}_2(\cdot)\) and a ReLU activation function:
\begin{equation}
\text{Path}_2(X) = \text{ReLU}\big(\text{BN}_2(W_2 \ast X + b_2)\big),
\end{equation}
where the bias term \( b_2 \in \mathbb{R}^{C_{\text{out}} / 2} \). The output tensor from this path is \( \text{Path}_2(X) \in \mathbb{R}^{B \times (C_{\text{out}} / 2) \times D' \times H' \times W'} \).

The outputs of the two paths are concatenated together along the channel dimension and fused through a \( 3 \times 3 \times 3 \) convolution with kernel \( W_{\text{merge}} \in \mathbb{R}^{C_{\text{out}} \times C_{\text{out}} \times 3 \times 3 \times 3} \), 
and a batch normalization layer \(\text{BN}_{\text{merge}}(\cdot)\):
\begin{equation}
Z = \text{BN}_{\text{merge}}\big(W_{\text{merge}} \ast \text{Concat}(\text{Path}_1(X), \text{Path}_2(X)) + b_{\text{merge}}\big),
\end{equation}
where \( b_{\text{merge}} \in \mathbb{R}^{C_{\text{out}}} \) is the bias term, \( \text{Concat}(\cdot) \) denotes the concatenation operation.

After fusing the features, the SE block \cite{Hu_2018_CVPR} is integrated to perform channel-wise attention. 
The SE block is a neural network module designed to enhance a network's representational capacity by modelling interdependencies between the channels of convolutional features. This block is incorporated into the present network to re-weight the feature maps of each channel during the processing of GPR segments. 
By applying attention mechanisms to channel dimensions, the SE block enables feature recalibration and allows the network to selectively emphasize informative features.

In the SE block, first, global average pooling is used to aggregate spatial information for each channel of the feature map tensor \( Z \in \mathbb{R}^{B \times C_{\text{out}} \times D' \times H' \times W'} \), where \( B \) is the batch size, \( C_{\text{out}} \) is the number of output channels, and \( D', H', W' \) denote the depth, height, and width of the feature maps, respectively. The pooled channel-wise descriptors are computed as:
\begin{equation}
S_c = \frac{1}{D' H' W'} \sum_{d=1}^{D'} \sum_{h=1}^{H'} \sum_{w=1}^{W'} Z_{bcdhw},
\end{equation}
producing a channel descriptor \( S_c \in \mathbb{R}^{B \times C_{\text{out}}} \). Two fully connected layers with a ReLU activation and a Sigmoid function are then applied to compute the channel-wise attention weights:
\begin{equation}
\alpha_c = \sigma\big(W_2^{\text{SE}} \, \text{ReLU}(W_1^{\text{SE}} S + b_1^{\text{SE}}) + b_2^{\text{SE}}\big),
\end{equation}
where \( W_1^{\text{SE}} \in \mathbb{R}^{r \times C_{\text{out}}} \) and \( W_2^{\text{SE}} \in \mathbb{R}^{C_{\text{out}} \times r} \) are the weight matrices of the two FC layers, and \( b_1^{\text{SE}} \in \mathbb{R}^r \), \( b_2^{\text{SE}} \in \mathbb{R}^{C_{\text{out}}} \) are the corresponding bias terms, and \( r \) is the reduction ratio. The function \( \sigma(\cdot) \) denotes the Sigmoid activation.
Finally, the learned attention weights \( \alpha_c \) are applied to recalibrate the original feature maps through channel-wise multiplication:
\begin{equation}
Z_c^{\text{SE}} = \alpha_c \cdot Z_c, 
\end{equation}
where \( Z_c^{\text{SE}} \) is the recalibrated feature map for the \( c \)-th channel of feature map \( Z_c\).

A shortcut connection is then added to preserve the input information. 
A \( 1 \times 1 \times 1 \) convolution is used to match the dimensions of features:
\begin{equation}
\text{Shortcut}(X) = \text{BN}_{\text{shortcut}}\big(W_{\text{shortcut}} \ast X + b_{\text{shortcut}}\big),
\end{equation}
where \(\text{BN}_{\text{merge}}(\cdot)\) is the batch normalization layer, \( W_{\text{shortcut}} \in \mathbb{R}^{C_{\text{out}} \times C_{\text{in}} \times 1 \times 1 \times 1} \) and \( b_{\text{shortcut}} \in \mathbb{R}^{C_{\text{out}}} \) are the weight tensor of the \(1 \times 1 \times 1\) convolution and the corresponding bias term. 
The final output is computed as:
\begin{equation}
Y = \text{ReLU}(Z_c^{\text{SE}} + \text{Shortcut}(X)).
\end{equation}

The output tensor \( Y \in \mathbb{R}^{B \times C_{\text{out}} \times D' \times H' \times W'} \) represents the recalibrated feature maps with enhanced representational capacity.

\subsubsection{Depthwise feature attention module}
The depthwise feature attention module is designed to selectively emphasize informative features along the depth dimension of a 3D input tensor. 
This attention mechanism operates independently for each depth segment and is implemented using a sequence of neural network layers.

Let the input tensor be \( X \in \mathbb{R}^{B \times C \times D \times H \times W} \), following the same indexing as above where \( B \), \( C \), \( D \), \( H \), and \( W \) represent the batch size, number of channels, depth, height, and width, respectively. 
The module first applies an adaptive average pooling operation across the spatial dimensions \( H \) and \( W \), retaining the depth dimension \( D \):
\begin{equation}
Y = \frac{1}{HW} \sum_{h=1}^{H} \sum_{w=1}^{W} X_{bcdhw}
\end{equation}
resulting in a tensor \( Y \) of dimensions \( B \times C \times D \times 1 \times 1 \).

Subsequently, two \( 1 \times 1 \times 1 \) convolutional layers process \( Y \). The first convolution applies a transformation with kernel \( W_1^{\text{DA}} \in \mathbb{R}^{C \times C \times 1 \times 1 \times 1} \) and bias \( b_1^{\text{DA}} \in \mathbb{R}^{C}\), followed by a ReLU activation:
\begin{equation}
Z^{(1)} = \text{ReLU}(W_1^{\text{DA}} \ast Y + b_1^{\text{DA}}).
\end{equation}

The second convolution applies a transformation with kernel \( W_2 \) and bias \( b_2 \), followed by a sigmoid activation, reducing the channel dimension to 1:
\begin{equation}
Z^{(2)} = \text{Sigmoid}(W_2^{\text{DA}} \ast Z^{(1)} + b_2^{\text{DA}}),
\end{equation}
where \( W_2^{\text{DA}} \in \mathbb{R}^{1 \times C \times 1 \times 1 \times 1} \) is the weight matrix and \( b_2^{\text{DA}} \in \mathbb{R}^{1} \) is the bias term, producing output \( Z^{(2)} \) of dimensions \( B \times 1 \times D \times 1 \times 1 \).

These scores are normalized using a softmax function across the depth dimension \( D \):
\begin{equation}
A_{bd} = \frac{e^{Z^{(2)}_{bd}}}{\sum_{k=1}^D e^{Z^{(2)}_{bk}}},
\end{equation}
where \( (A_{bd} \in \mathbb{R}^{B \times D} \) denotes the attention scores, with one score per depth segment \( d \) for each sample in the batch \( b \). This ensures that each depth segment receives a weighted emphasis based on its computed significance.

Finally, the attention scores are expanded back to the dimensions of the input tensor and applied element-wise:
\begin{equation}
O_{bcdhw} = A_{bd} \times X_{bcdhw}, 
\end{equation}
where \( O \in \mathbb{R}^{B \times C \times D \times H \times W} \) is the output tensor \( X \in \mathbb{R}^{B \times C \times D \times H \times W} \), which retains the same dimensions as the input \( X \) but includes features along the depth dimension according to their learned significance.
By reweighting the input tensor, this module emphasizes the most informative regions across the depth segments.


\section{Experiments and Results}
\label{sec_result}

\subsection{Implementation details}

\begin{figure}[H]
         \begin{subfigure}{0.4\textwidth}
	\centering
  	\includegraphics[width=\linewidth]{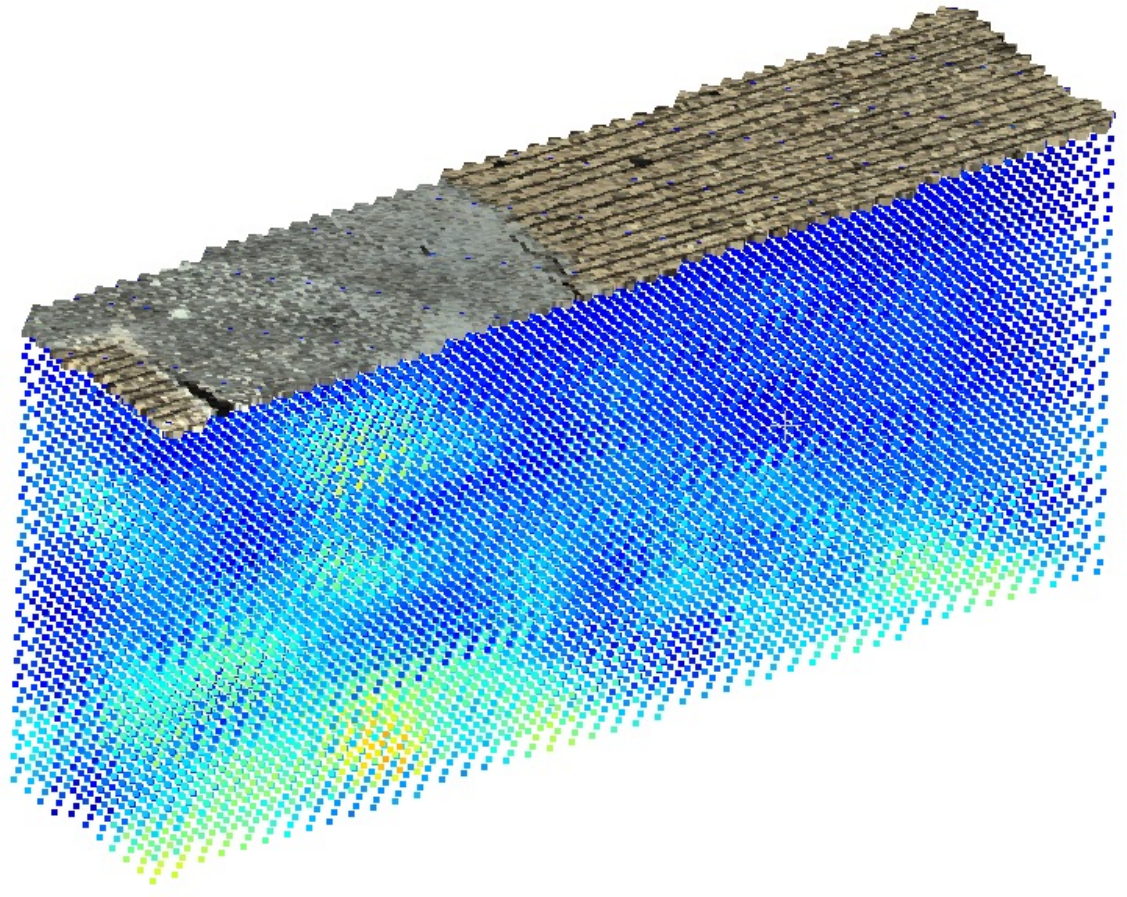}
  	\caption{Patch segment}
        \end{subfigure}     
        \begin{subfigure}{0.4\textwidth}
	\centering
  	\includegraphics[width=\linewidth]{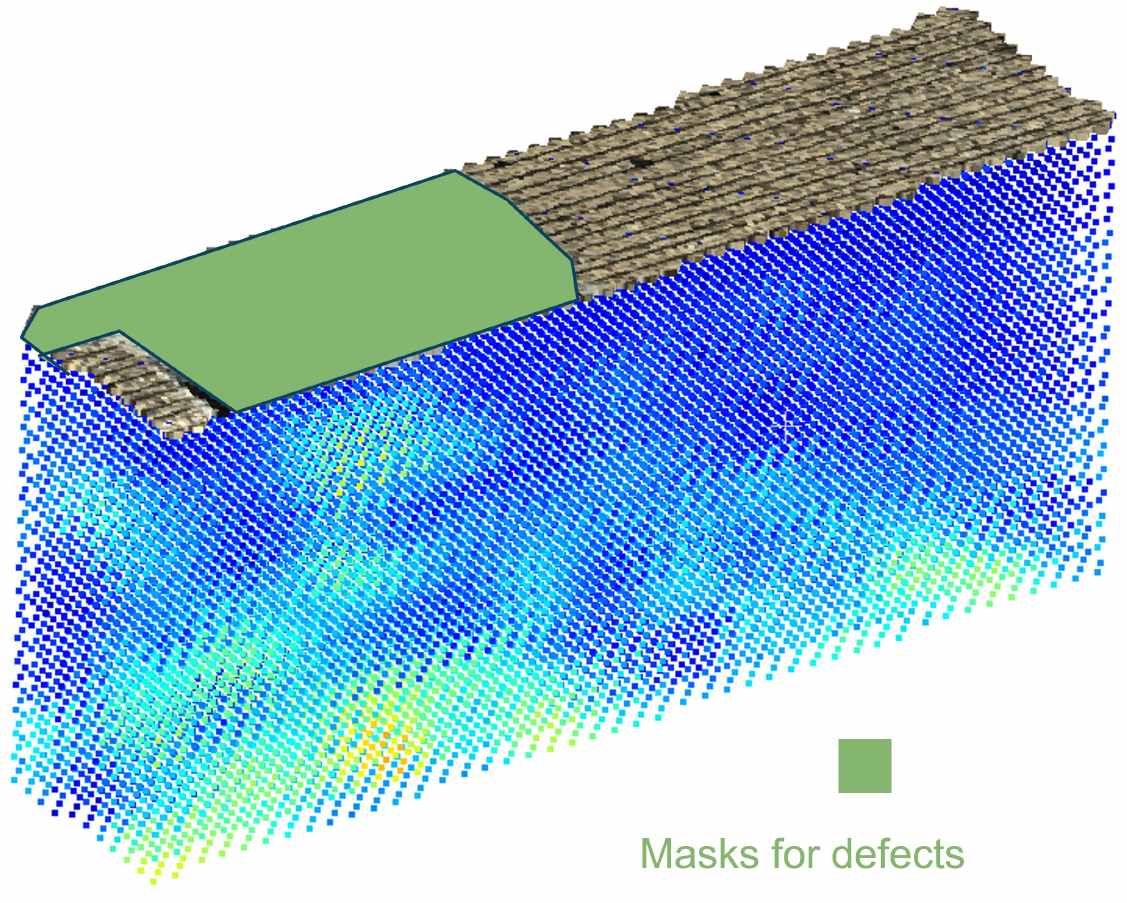}
  	\caption{Patch segment with annotation}
        \end{subfigure}    
        
        \begin{subfigure}{0.4\textwidth}
	\centering
  	\includegraphics[width=\linewidth]{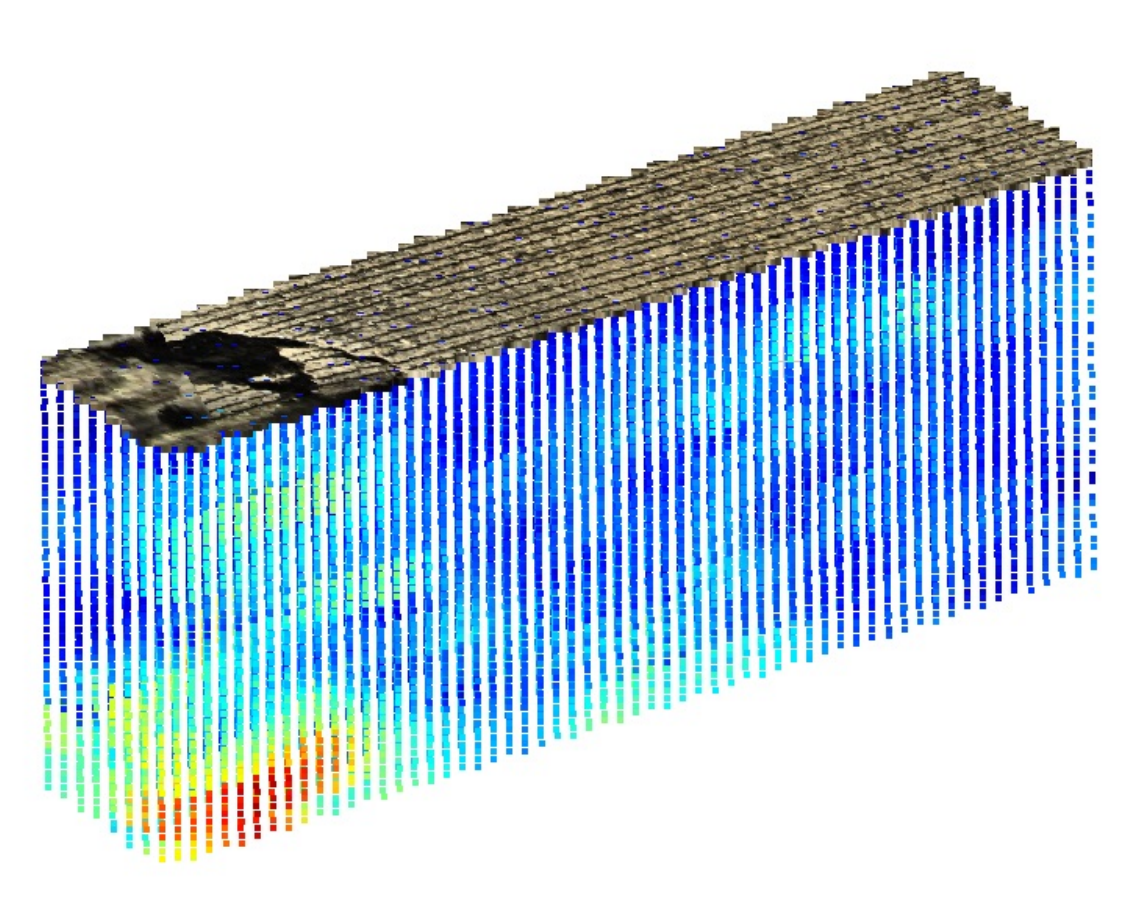}
  	\caption{Crack segment}
        \end{subfigure}        
      \begin{subfigure}{0.4\textwidth}
	\centering
  	\includegraphics[width=\linewidth]{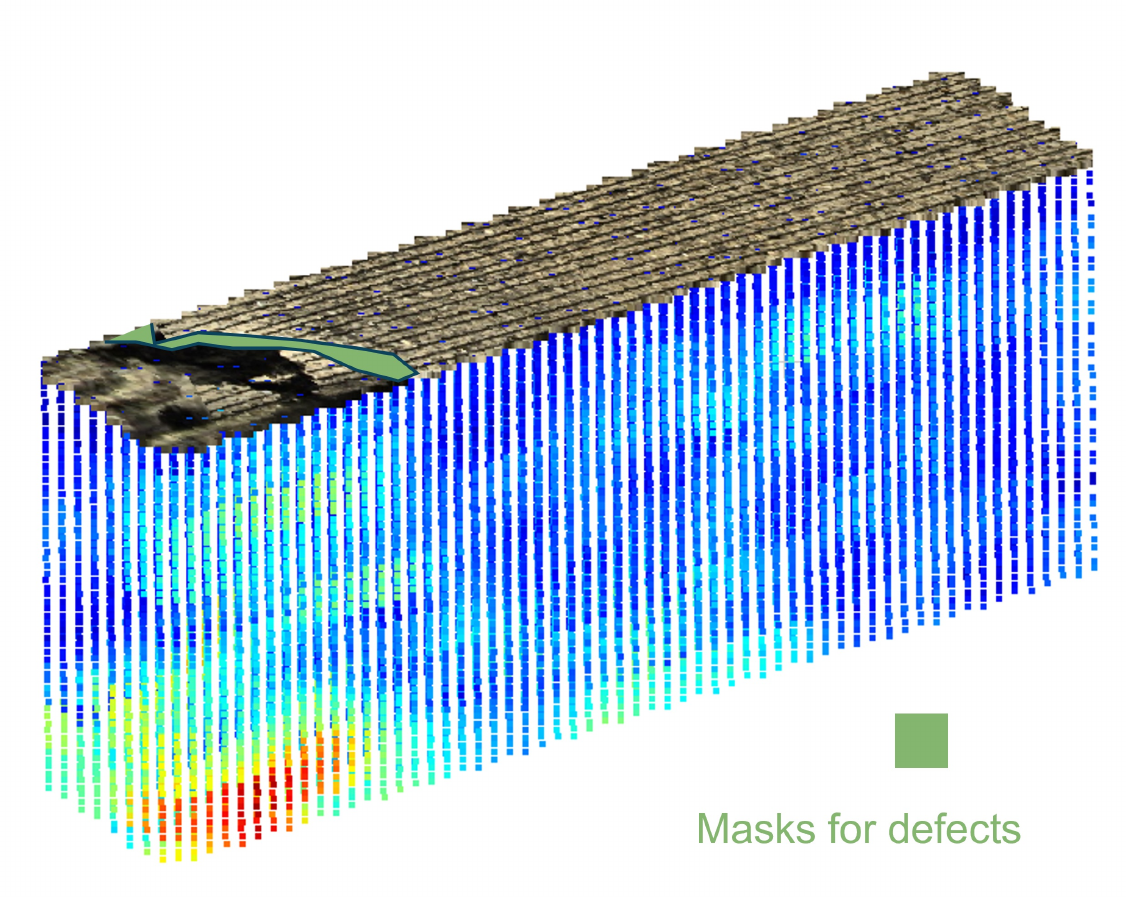}
  	\caption{Crack segment with annotation}
        \end{subfigure}   
        \caption{RGB-GPR segments with annotations}
        \label{fig:defectmask}  
\end{figure}

The approach of constructing the GPR-RGB dataset is detailed in Section \ref{subsec:dataset}.  
Two types of pavement defects in the original CamHighway dataset are extracted on the pavement images: \emph{patches} and \emph{cracks}. These annotations were then transferred to the corresponding GPR-RGB data segments. Two examples of GPR segments with patch and crack defects are illustrated in Figure \ref{fig:defectmask}. Specifically, if a defect is visible in the RGB image, the associated GPR segment is labelled with the same defect type.
These labels are treated as RGB-guided proxy labels for learning correlations between surface-visible defects and GPR response patterns, rather than as independently verified ground truth for subsurface distress. Therefore, the proposed annotation strategy is applicable mainly to defects with visible surface manifestations and does not provide labels for hidden internal distresses that are not observable from images. No additional field spot-checks, coring, or independent physical validation of the model predictions were conducted, as the data were collected from a highway section under continuous operation. 
The total number of generated segments, along with the sample distribution across the two defect classes and the train-test split, is summarised in Table \ref{table:number}.

\begin{table}[H]
\caption{Sample number of different classes in prepared dataset}
\label{table:number}
\centering
\small
\renewcommand{\arraystretch}{1.25}
\begin{tabular}{ lllll}
\hline\hline
Class & Training Set & Test Set & Total\\
\hline
Patch & 652&220 & 872\\
Crack	& 633 &140&773\\
No defect &	800&300&1100\\
Total & 2085 &660&2745	\\ 
\hline\hline
\end{tabular}
\end{table}

The dataset preparation and the neural networks were implemented in Python. 
The experiments were conducted on the customized dataset described above on an NVIDIA A100 GPU with 80 GB of memory.



\subsection{Results and outcome analysis}
\label{subsec:results}
\begin{figure}[H]
         \begin{subfigure}{0.4\textwidth}
	\centering
  	\includegraphics[width=\linewidth]{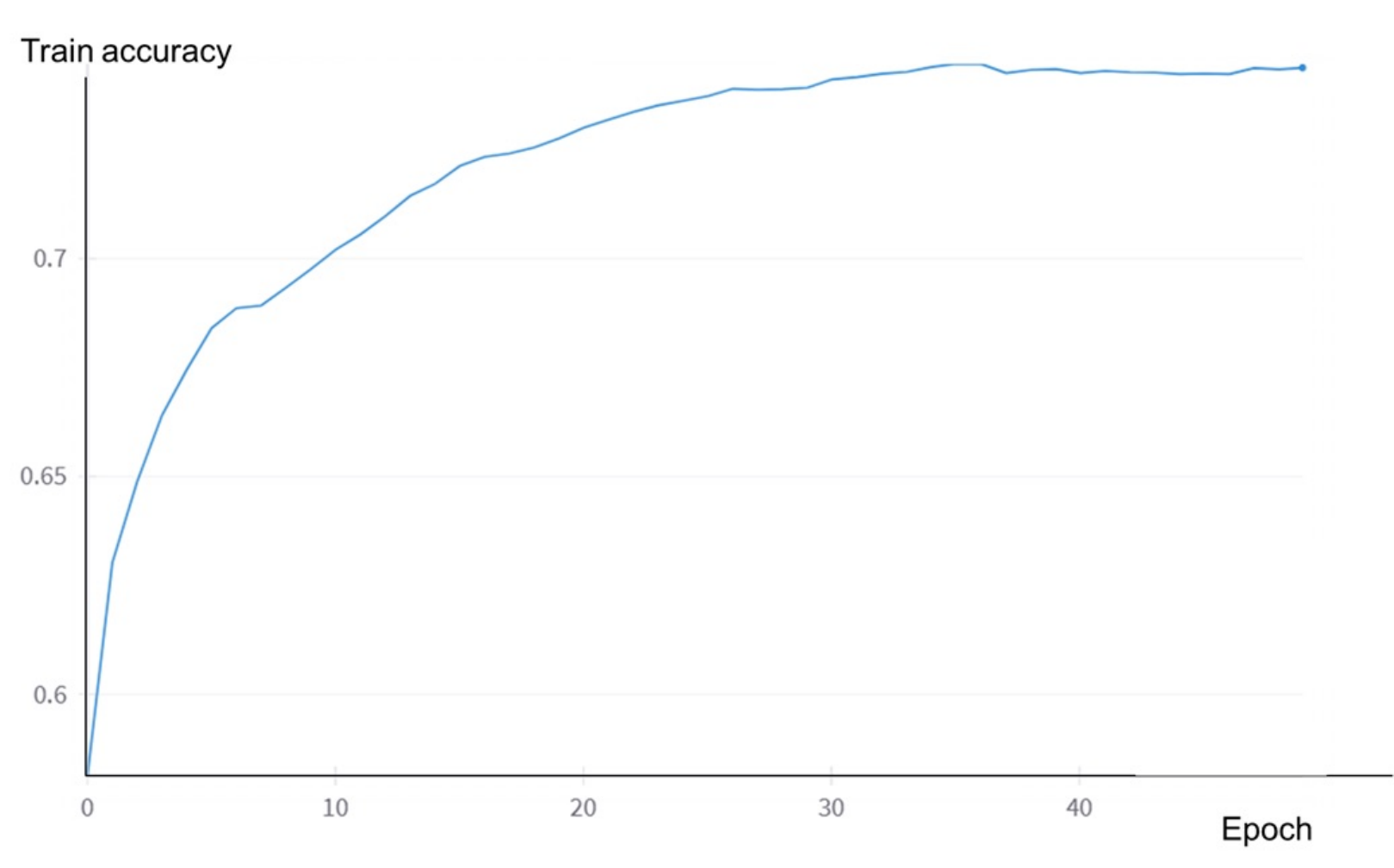}
  	\caption{Training accuracy curve for patch classification}
        \end{subfigure}     
        \begin{subfigure}{0.4\textwidth}
	\centering
  	\includegraphics[width=\linewidth]{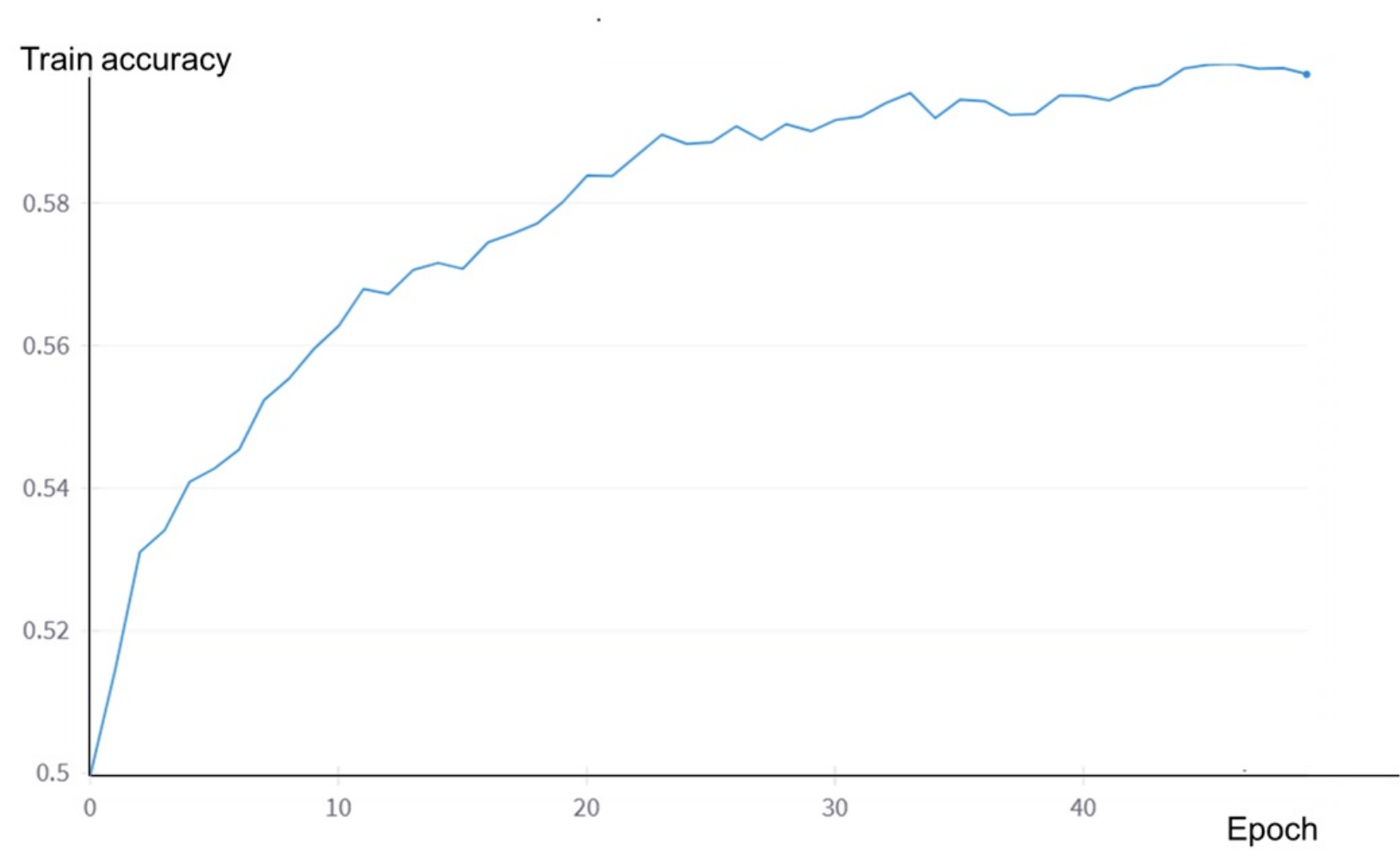}
  	\caption{Training accuracy curve for crack classification}
        \end{subfigure}    
         
        \caption{Training curves for proposed network}
        \label{fig:trainingcurve}  
\end{figure}

To ensure a fair performance comparison, each experimental model is trained and tested four times, and the average of the resulting evaluation metrics is reported. All tests are formulated as binary classification problems. Specifically, these models are trained on data labelled as "patch vs. normal" and "crack vs. normal". 
To evaluate the performance of the proposed network, three neural networks are implemented: a) ResNet-18 \cite{he2016deep} with and without pre-training on the 2D ImageNet dataset \cite{deng2009imagenet}. Given the significant differences between natural images and GPR data, we hypothesise that pre-training may not yield noticeable performance improvements. b) A 3D CNN with the same number of layers and convolutional kernel sizes as the proposed architecture, but without additional enhancements. c) The proposed network with residual blocks, mixed convolutional kernel sizes, and both channel-wise and depth-wise feature attention mechanisms. 
To ensure a fair evaluation, all networks have the same configuration: a batch size of 2, a maximum of 50 epochs. The AdamW optimiser is employed, which improves generalisation by using weight decay from gradient updates. A step learning rate scheduler was applied with a step size of 10 and a decay factor $\gamma$ of 0.1. To address class imbalance in the training data, a weighted binary cross-entropy loss function $\mathcal{L}$ was used to assign higher penalties to underrepresented classes during model training:
\begin{equation}
\mathcal{L} = - \left( w_1 \, y \log(p) + w_0 \, (1 - y) \log(1 - p) \right),
\end{equation}
where \( y \in \{0, 1\} \) is the ground truth label, \( p \in [0, 1] \) is the predicted probability for the positive class, and \( w_1, w_0 \in \mathbb{R}^+ \) are the weights for the positive and negative classes, respectively. Figure \ref{fig:trainingcurve} shows the training accuracy curves of the proposed network for both the patch and crack classification tasks.

We use the following standard metrics to evaluate the performance of the networks: accuracy, precision, recall, and F1 score. 
These metrics are defined by the following equations:
\begin{equation} \text{Accuracy} = \frac{\text{TP} + \text{TN}}{\text{TP}+ \text{TN} + \text{FP} + \text{FN}}, \label{equ:acc} 
\end{equation}
\begin{equation} 
\text{Precision} = \frac{\text{TP}}{\text{FP} + \text{TP}}, \label{equ:prec} 
\end{equation} 
\begin{equation} 
\text{Recall} = \frac{\text{TP}}{\text{TP} + \text{FN}}, \label{equ:recal} 
\end{equation} 
\begin{equation} 
\text{F1 Score} = \frac{2 \times \text{Precision} \times \text{Recall}}{\text{Precision} + \text{Recall}}, \label{equ:f1} 
\end{equation} 
where TP, TN, FP, and FN denote true positive, true negative, false positive, and false negative, respectively. 
The quantitative evaluation of different networks on the test set is presented in Table \ref{table:deep1}.
The reported values for precision, recall, and F1-score are macro-averaged across two classes, which averages these metrics for the two classes to assess the overall performance, providing a comprehensive assessment of each model's effectiveness.

\begin{table}[H]
\caption{Prediction performance of various networks on test set \((\%)\)}
\label{table:deep1}
\centering
\small
\renewcommand{\arraystretch}{1.25}
\begin{tabular}{  llllll }
\hline\hline
Defect & Model& Accuracy & Precision & Recall & F1 \\
\hline
Patch & Resnet18&59.6	&58.9	&59.0	&58.9 \\
& Resnet18 (pretrain)&59.3 &	59.4	&59.4&	58.8 \\
&3D Conv& 67.6&	66.9&	65.8&	65.9 \\
& Proposed network&\textbf{69.0} &	\textbf{68.2}&	\textbf{68.0}&	\textbf{68.1}\\
\hline
Crack & Resnet18& 54.9&	52.4&	52.7&	51.8\\
& Resnet18 (pretrain) &54.0&	51.4&	51.6&	50.7\\
&3D CNN& 68.6&	64.3&	\textbf{64.1}&	\textbf{64.1} \\
& Proposed network&\textbf{69.4}&	\textbf{64.6}&	64.0&	\textbf{64.1}\\
\hline\hline
\end{tabular}

\end{table}

For patch defect detection, the ResNet-18 network achieved accuracies of 59.6\% and 59.3\% for models without and with pre-training on the ImageNet dataset \cite{deng2009imagenet}, respectively. 
Each GPR segment is represented as a tensor of size \(100 \times 25 \times 20 \times 3\), where the dimensions correspond to the longitudinal direction, lateral direction, depth, and RGB colour channels, respectively. Since the original ResNet-18 model is designed to process standard 2D images with three channels, the GPR segments were reformatted to \(100 \times 25 \times 60\) by flattening the depth and colour dimensions into a single channel dimension, whilst preserving the spatial resolution. 
To support this input format, the first convolutional layer of the ResNet-18 architecture was modified to accommodate the transformed GPR segments.
As shown in Table \ref{table:deep1}, pre-training with weights derived from conventional 2D image datasets does not improve performance metrics. Although transfer learning typically enhances object detection accuracy in different applications in the Architecture, Engineering, and Construction industry \cite{PAN2022104375,pan2022automatic}, the features learned from purely 2D images do not effectively transfer to 3D GPR segment data. This limitation likely arises from the fundamental differences in dimensionality and data characteristics between 2D imagery and 3D GPR representations. 

For the patch classification task, compared to the ResNet-18 model, which processes GPR data as 2D data, the 3D-CNN model provides significantly improved performance, increasing the accuracy from 59.6\% to 67.6\%, precision from 58.9\% to 66.9\%, recall from 59.0\% to 65.8\%, and F1 score from 58.9\% to 65.9\%. 
This improvement is expected since the 3D convolutional layers are adept at extracting features from 3D spaces. 
Given that GPR segments inherently possess a 3D structure, the ability to extract features along the depth dimension is crucial for accurately predicting the classes of these segments.

Building on the enhancements provided by the 3D convolutional layers, the proposed network incorporates residual connections, mixed convolutional kernel sizes, and attention mechanisms, resulting in further improvements across the various performance metrics. 
Specifically, accuracy increases from 67.6\% to 69.0\%, precision from 66.9\% to 68.2\%, recall from 65.8\% to 68.0\%, and F1-score from 65.9\% to 68.1\%. 
These results demonstrate that the proposed model provides meaningful and consistent improvements over existing baselines. Its effectiveness is particularly notable given the challenges posed by real-world GPR data, which are often noisy and ambiguous due to the complexity of road environments.

For crack defect detection, similar patterns in prediction results were observed. 
The ResNet-18 models achieved lower accuracies of 54.0\% and 54.9\% with and without pre-training on ImageNet, respectively.
Once again, the 3D-CNN model demonstrated a notable improvement, increasing the accuracy to 69.1\%. 
However, the present network provides slightly improved performance, achieving the highest predicted accuracy of 69.6\%.

The performance improvement from the 3D CNN to the proposed network is modest compared to the gain from ResNet-18 to the 3D CNN. To assess its significance, we conducted paired t-tests across four training runs using identical data splits. This test is suitable here as it evaluates differences under matched conditions. For patch classification, the p-value is 0.0272, indicating a statistically significant improvement ( \(p < 0.05\)) at the 95\% confidence level. For crack classification, the p-value is 0.0938, suggesting moderate evidence of improved performance at the 90\% confidence level, though not significant at 95\%. In summary, the proposed network significantly outperforms the 3D CNN in patch classification and shows a consistent, albeit weaker, advantage in crack classification.

\begin{figure}[H]
\centering
         \begin{subfigure}{0.48\textwidth}
            \centering
            \includegraphics[width=\linewidth]{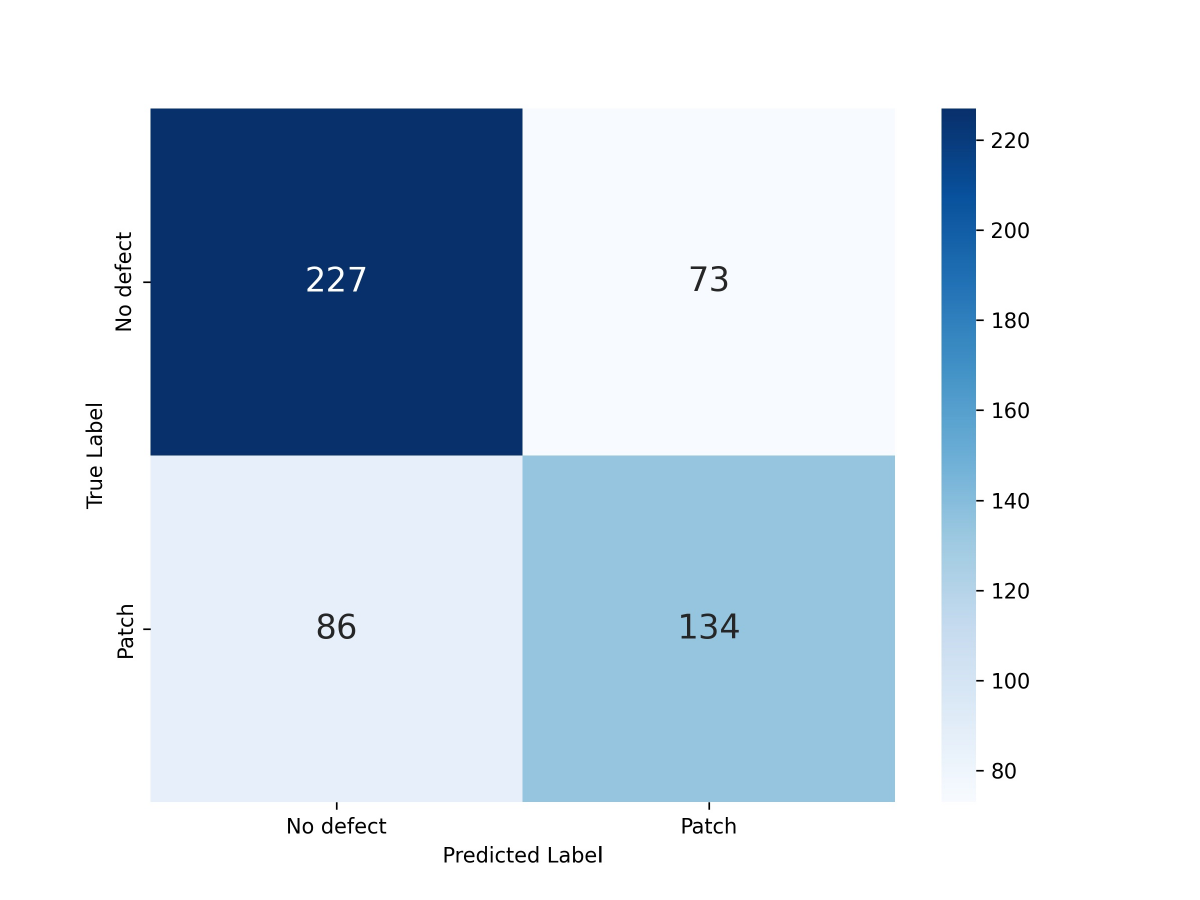}
            \caption{Confusion matrix for patch classification}
            \label{fig:patchconf}
        \end{subfigure}
	    \begin{subfigure}{0.48\textwidth}
            \centering
            \includegraphics[width=\linewidth]{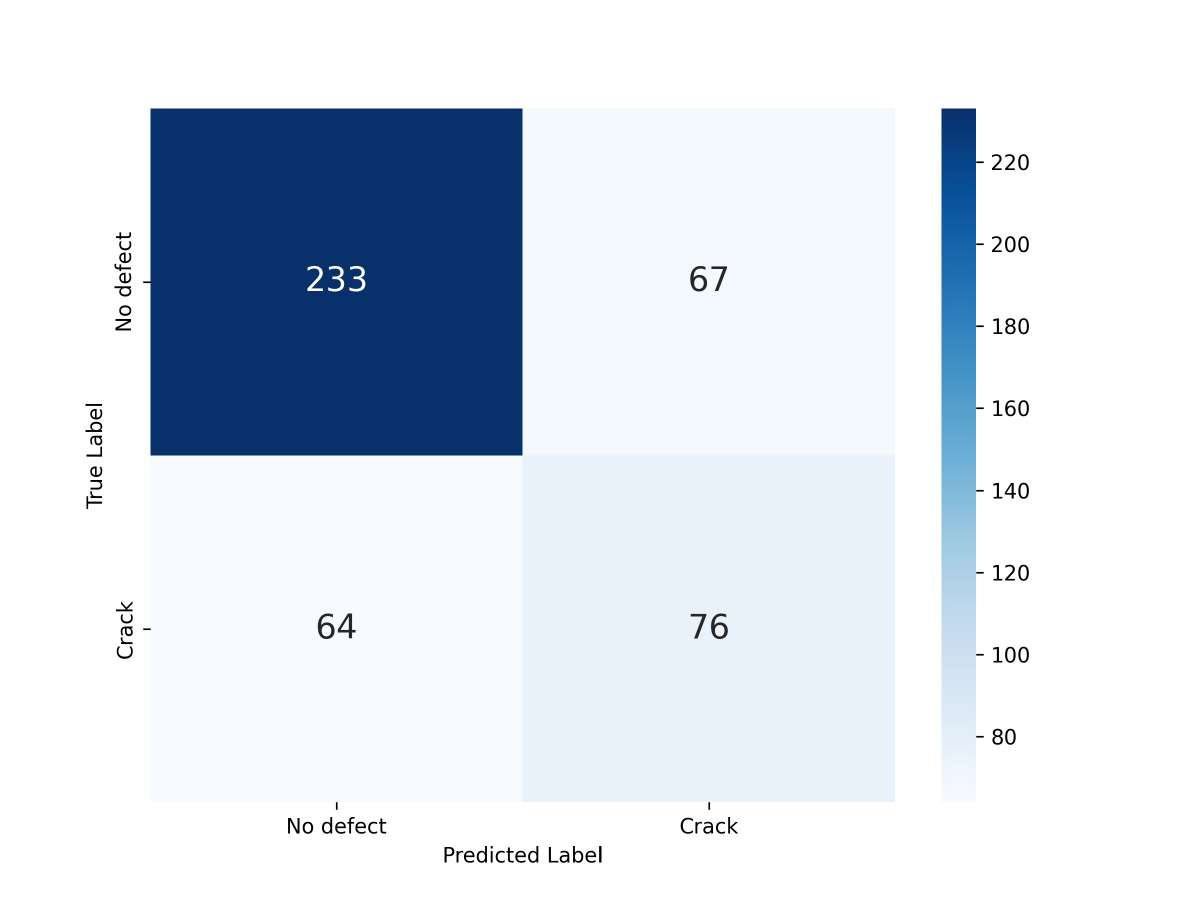}
            \caption{Confusion matrix for crack classification}
            \label{fig:crackconf}
        \end{subfigure}  
        \caption{Confusion matrix for patch and crack classification tasks}
        \label{fig:confusion}
\end{figure}

Despite achieving similar overall accuracies of approximately 70\%, the model's performance for two tasks, patch defect detection and crack defect detection, varies when examining the patch and crack classes in detail.
This variation is detailed in Table \ref{table:deep_detailed}, where class-wise precisions, recalls, and F1 scores are presented.
It is evident that the model's ability to detect patches is substantially better than identifying cracks, a finding that is supported by the confusion matrices presented in Figure \ref{fig:confusion}.

\begin{table}[H]
\caption{Classwise comparison for patch and crack classification \((\%)\)}
\label{table:deep_detailed}
\centering
\small
\renewcommand{\arraystretch}{1.25}
\begin{tabular}{  llllll }
\hline\hline
Defect type & Class  & Precision  & Recall & F1 \\
\hline
Patch & Normal &	72.5&	75.7&	74.1 \\
                & Patch &	64.7&	60.9&	62.8 \\
\cline{2-5}
                & \textit{Average} &  68.6&	68.3&	68.4 \\
\hline
Crack & Normal &	78.5&	77.7&	78.1 \\
                & Crack &	53.2&	54.3&	53.7 \\
\cline{2-5}
                & \textit{Average} &   65.8&	66.0&	65.9 \\
\hline\hline
\end{tabular}

\end{table}

This discrepancy can be attributed to the following factors:
a) The causal link between GPR anomalies and surface defects is likely to differ between patches and cracks. For instance, when the GPR anomaly reflects increased subsurface water content, the presence of a patch is more likely to correlate strongly. Patches often result from prior surface damage that permits significant water ingress into the underlying material. In contrast, cracks are typically narrow and allow limited water penetration, leading to subtler subsurface changes that produce weaker GPR responses.
b) The model's ability to detect patches and cracks is different. Patch defects are generally larger than cracks, making them more detectable with the employed techniques. Furthermore, patch defects tend to have more regular shapes as they are human-generated during repair processes, whereas crack defects occur more randomly and result from various causes, presenting diverse manifestations across environments. 
c) The spatial resolution of the GPR data in this study is approximately 5 cm, which leads to information loss for finer-scale features. This limitation particularly affects the detection of cracks, which are often narrow and fall below the resolution threshold.
d) The data utilized in this study are collected from real-world environments and are highly noisy. This background noise likely obscures subtle patterns associated with cracks, making them harder to distinguish from the surrounding signal.

It should also be noted that the classification labels used in this study are derived from surface-visible image defects. Therefore, the reported results evaluate the ability of the model to recognise GPR patterns associated with these visible defects, rather than its ability to identify all forms of subsurface distress. In particular, defects such as voids or delamination that do not show surface manifestations are outside the scope of the present dataset and model.

The experiments in this study were formulated as two binary classification tasks, namely patch versus normal and crack versus normal, because the primary objective was to evaluate whether GPR segments associated with each type of surface-visible defect could be distinguished from normal pavement segments. This formulation is also consistent with preliminary screening workflows in road inspection, where candidate abnormal segments are first identified for further review. A three-class setting involving normal, patch, and crack classes was also considered. However, preliminary results indicated that the current model and dataset have difficulty distinguishing crack and patch segments from each other. This suggests that the GPR response patterns associated with RGB-labelled cracks and patches are not always clearly separable in the current dataset, likely due to the weak-label nature of RGB-guided annotation, limited GPR resolution, and noisy real-world highway data. 

From the perspective of practical road maintenance, the proposed model should be regarded as a preliminary screening or decision-support tool rather than a standalone system for determining maintenance actions. The relatively low F1-score for crack detection and the moderate recall for patch detection indicate that false positives and false negatives remain substantial. False positives may increase unnecessary follow-up inspections, while false negatives may cause relevant defects to be missed, which is more critical in safety-related scenarios. Therefore, the model outputs should be used to select candidate road segments for further inspection, rather than directly starting maintenance actions. Further improvements in data quality, dataset size, model robustness, and independent field validation are needed before operational deployment.

\subsection{Ablation study}
\label{sec:ablation}

The network architecture proposed in Section \ref{subsec:network} incorporates 3D convolutional layers enhanced with residual connections, and utilizes mixed convolutional kernel sizes along with depthwise and channelwise attention mechanisms to optimize feature extraction from GPR segments. 
Expanding on the standard 3D CNN framework, various configurations are explored: networks with individual integrations of residual connections, depth attention, channel attention, and mixed kernel sizes, as well as the proposed network that combines all these techniques. 
The comparative performance of these configurations for patch classification is detailed in Table \ref{table:deep2}.

\begin{table}[H]
\caption{Ablation study of individual techniques on patch classification \((\%)\)}
\label{table:deep2}
\centering
\small
\renewcommand{\arraystretch}{1.25}
\begin{tabular}{  lllll }
\hline\hline
Model& Accuracy  & Precision  & Recall & F1 \\
\hline
3DCNN&67.6&	66.9&	65.8&	65.9 \\
Residual block&66.8 &	66.0&	65.9&	66.0 \\
Depth attention& 68.3 &	67.4&	67.2&	67.3 \\
Channel attention&67.5 &	66.9&	66.7&	66.6\\ 
Mixed kernel size&66.7 &	65.9&	65.8&	65.8\\
Proposed network&\textbf{69.0} &	\textbf{68.2}&	\textbf{68.0}&	\textbf{68.1}\\
\hline\hline
\end{tabular}

\end{table}


Similar to the experiments outlined in Table \ref{table:deep1}, all networks were trained and tested five times, with the average values from these trials reported in Table \ref{table:deep2}. The ablation results show that the contribution of each architectural component is not uniformly positive when evaluated independently. Compared with the baseline 3D CNN, the depth attention mechanism improves all four evaluation metrics, increasing the accuracy from 67.6\% to 68.3\% and the F1-score from 65.9\% to 67.3\%. The channel attention mechanism provides modest improvements in recall and F1-score, although its accuracy remains similar to the baseline. The residual block and mixed-kernel module do not improve accuracy when used independently.

These results suggest that the proposed architecture benefits from the synergistic interaction among its components rather than from a simple cumulative improvement of each individual module. Although residual connections and mixed kernels do not improve performance in isolation, they may support feature propagation and multi-scale feature representation when combined with depth-wise and channel-wise attention.

\subsection{Influence of GPR input layers}
\label{sec:chosen_gpr_layer}

As described in Section \ref{subsec:dataset}, the GPR data utilized in this study contain 20 layers, extending from the pavement surface downward into the subsurface, with a 5 cm interval between adjacent layers.
Table \ref{table:deep3} compares the performance of the proposed network in classifying patch defects across four configurations of input layers: 5, 10, 15, and 20 layers from the top.
It can be observed that configurations with a larger number of input layers achieve better scores, with the 5-layer input performing the worst and the 20-layer input demonstrating the best prediction results. However, this trend should be interpreted with caution. Patch defects usually extend only to a limited depth. Therefore, the improved classification performance obtained by including more GPR slices should not be interpreted as evidence that the physical patch defect extends to the deepest layers included in the model input.

A more reasonable interpretation is that the additional GPR slices provide contextual information that helps the model distinguish patched and non-patched segments. The upper GPR slices are expected to contain the most direct information related to patch materials, particularly when the dielectric properties of the patch material differ from those of the neighbouring materials. Deeper slices may still contribute indirectly by capturing broader GPR signal characteristics, such as reflection patterns, attenuation behaviour, or moisture-related dielectric changes. In this sense, the benefit of using more slices reflects the contribution of 3D contextual information rather than the physical depth extent of the patch defect itself.

\begin{table}[H]
\caption{Results of evaluating different number of input layers on proposed model for patch classification (\%)}
\label{table:deep3}
\centering
\small
\renewcommand{\arraystretch}{1.25}
\begin{tabular}{ lllll }
\hline\hline
 Layer Number& Accuracy & Precision & Recall & F1 \\
\hline
  5 &67.5&	66.8&	65.9&	65.9 \\
 10& 68.1&	67.3&	67.0&	67.1 \\
  15&68.5&	67.7&	67.5&	67.5\\
   20&\textbf{69.0} &	\textbf{68.2}&	\textbf{68.0}&	\textbf{68.1}\\
\hline\hline
\end{tabular}

\end{table}

However, no such trend is observed for crack defects within the dataset. 
From Table \ref{table:deep4}, utilizing only the top 5 layers results in the poorest prediction outcomes.
However, adding more layers did not consistently improve all metrics.
This variability can be attributed to the irregular nature of cracks compared to patches, which are more uniformly constructed during pavement maintenance. 
Cracks vary significantly based on different factors such as their causes and environmental conditions, presenting diverse properties and appearances.
Deeper layers in the GPR data may decrease model performance when they are irrelevant to the defect itself. 
However, the low spatial resolution and high noise levels in the collected GPR data, combined with the irregular and small-scale nature of cracks, make crack detection particularly challenging. These issues also affect the ability to analyse and identify performance trends when varying the number of input GPR layers along the depth dimension.

\begin{table}[H]
\caption{Results of evaluating different number of input layers on proposed model for crack classification (\%)}
\centering
\small
\renewcommand{\arraystretch}{1.25}
\begin{tabular}{  lllll }
\hline\hline
 Layer Number& Accuracy & Precision  & Recall  & F1  \\
\hline
  5 &66.1&	60.6&	55.5&	53.4 \\
 10& \textbf{70.6}&	65.8&	64.1&	64.4 \\ 

  15&70.0&	65.5&	65.5&	65.4\\
   20&69.0 &	\textbf{68.2}&	\textbf{68.0}&	\textbf{68.1}\\
\hline\hline
\end{tabular}

\label{table:deep4}
\end{table}

\section{Conclusions}
\label{sec_conclusion}

This paper presented a novel pipeline for analysing 3D GPR data using RGB-guided annotation and deep learning. We introduced a cost-effective method to construct annotated 3D GPR datasets by aligning orthomosaic images with GPR scans, and designed a specialised 3D convolutional neural network enhanced with residual connections, attention mechanisms, and mixed kernel sizes to improve feature extraction. The proposed network was evaluated on patch and crack detection tasks, with experimental results demonstrating its better performance over baseline approaches.
This paper addresses the research gaps discussed and makes significant contributions to state-of-the-art GPR research in several key aspects:

a) This paper introduced a novel network architecture specifically designed to process 3D GPR data and classify two common types of pavement defects: patches and cracks.
The proposed network demonstrated meaningful and consistent improvements over existing baselines. 

b) A 3D GPR dataset collected in the real world on a section of concrete pavement on a section of highway in use (A14 in the UK) was presented. 
It enabled the comparative evaluation of the performance of various models on real-world data.
It is important to emphasize the value of a real-world dataset. The GPR data contained within are notably more complex, irregular, and noisy, due to numerous factors that cannot be modeled in a GPR simulator. 

c) This paper presented a novel approach involving the alignment of RGB-GPR data, using pavement RGB images as a reference for annotating GPR data. This approach enabled the generation of a large volume of labels by first labelling the images and then transferring these annotations to the corresponding GPR segments. 
Given that labelling an image is significantly faster and more cost-effective compared to directly labelling GPR data, which often requires techniques like coring or specialized human expertise, this approach offers a practical alternative for data annotation.
Furthermore, the annotated RGB-GPR segments provide possible future research directions, such as exploring deep learning techniques across these two data modalities. 


Practically, this study makes important contributions to the civil and transportation engineering domain by proposing a scalable and cost-effective approach for pavement defect detection using GPR data. The novel idea of using RGB images as a reference to label 3D GPR data addresses a major bottleneck in the field: the lack of real-world annotated datasets. It enables large-scale data generation under real-world conditions. With such datasets available, more advanced machine learning and deep learning models can be developed and applied to improve defect recognition performance. 

Furthermore, the integration of GPR and pavement image data improves both the efficiency and depth of pavement condition assessment, allowing not only for the detection of surface and subsurface defects but also for the analysis of underlying causes. This integrated insight is especially valuable for pavement engineers and supports intelligent platforms like road digital twins \cite{brilakis_pan_borrmann_mayer_rhein_vos_pettinato_wagner_2019,PAN2024105654}, as it supports more informed and proactive decision-making in maintenance planning and repair prioritisation, contributing to more effective and data-driven road asset management.


However, this study still has the following limitations: a) the RGB-guided annotation strategy provides proxy labels rather than physically verified ground truth for subsurface distress. 
This study assumes that defects visible in RGB images always correspond to distinguishable defect patterns in GPR data.
However, there is no guaranteed correlation that a defect on the pavement surface always corresponds to a subsurface anomaly, or vice versa. Accurately annotating these subsurface defects usually requires complementary approaches, such as coring or field checks. This limitation applies to the prediction results as well. The validation and testing phases were based on samples from the constructed dataset, without independent physical validation to verify the model predictions. Future work should include independent validation through coring, field spot-checks, or complementary non-destructive testing methods, where feasible, to further assess the reliability of the method in practical pavement maintenance.
b) The characteristics of GPR data can vary substantially across different environments, depending on subsurface material properties such as density, composition, and moisture content. For example, GPR signals interact differently with sandy soils compared to clayey soils due to differences in dielectric properties. These variations can change the appearance of anomaly patterns in the GPR data. As a result, the model trained on data collected from a specific environment may not generalise well to GPR data acquired under different ground conditions or in different geographical locations. 
Models trained on the dataset used in this study might not perform effectively on GPR data collected from other environments. 
c) The current model performance, especially for crack recognition, is not yet sufficient for standalone operational decision-making. In practical road maintenance, the model should be used as a screening tool to support further inspection. Further improvements in aspects like data quality, data resolution, and model robustness are needed before real operational deployment.

In future work, we aim to expand the GPR-RGB dataset to include a wider variety of pavement types, and additional geographical areas.
A more expansive dataset will enable the development of more comprehensive models capable of performing additional tasks.
Furthermore, we intend to include other types of pavement defects, such as rutting, bleeding, ravelling, and potholes, to enhance the dataset's applicability and support more generalised defect detection.

\section{Data Availability Statement}
Data that support the findings of this study are available from the corresponding author upon reasonable request.

\section{Acknowledgements}
This project has received funding from the European Union’s Horizon 2020 research and innovation programme under the Marie Skłodowska-Curie grant agreement No 101034337.

%
%
\bibliography{ascexmpl-new}

\end{document}